%% file: arxiv_v1.tex
\documentclass{arxiv}

\OneAndAHalfSpacedXI

\usepackage{natbib}
 \bibpunct[, ]{(}{)}{,}{a}{}{,}%
 \def\bibfont{\small}%
\usepackage{tikz}
\usetikzlibrary{positioning}
\usepackage{pgflibraryshapes}
\usepackage{times}
\usepackage{soul}
\usepackage{url}
\usepackage[hidelinks]{hyperref}
\usepackage[utf8]{inputenc}
\usepackage[small]{caption}
\usepackage{amsmath}
\usepackage{booktabs}
\usepackage{algorithm}
\newcommand{\ie}{\textit{i}.\textit{e}., }

\usepackage{amsmath,amssymb}
\usepackage{graphicx}
\usepackage{subcaption}
\usepackage{booktabs} % for professional tables
\usepackage{algpseudocode}
\usepackage{algorithm}
\usepackage{bbm}
\usepackage[utf8]{inputenc}
\usepackage[english]{babel}

\usepackage{hyperref}
\usepackage{cleveref}
\usepackage{mathtools}

\usepackage{graphicx}
\usepackage{subcaption}

\TheoremsNumberedThrough     % Preferred (Theorem 1, Lemma 1, Theorem 2)
\ECRepeatTheorems

\EquationsNumberedThrough    % Default: (1), (2), ...
\MANUSCRIPTNO{MS-0001-1922.65}

\begin{document}
%%%%%%%%%%%%%%%%

%\setcounter{page}{1}
%\DOI{10.1287/xxxx.0000.0000}%


\RUNTITLE{Learning Complementary Policies for Human-AI Teams}

% \TITLE{Bundling Information Goods of Decreasing Value}
\TITLE{Learning Complementary Policies for Human-AI Teams}

\ARTICLEAUTHORS{%
\AUTHOR{Ruijiang Gao}
\AFF{University of Texas at Dallas, \EMAIL{ruijiang.gao@utdallas.edu}} %, \URL{}}
\AUTHOR{Maytal Saar-Tsechansky}
\AFF{University of Texas at Austin, \EMAIL{maytal.saar-tsechansky@mccombs.utexas.edu}}
\AUTHOR{Maria De-Arteaga}
\AFF{ESADE Business School, \EMAIL{maria.dearteaga@esade.edu}}
} % end of the block

\ABSTRACT{%
 This paper tackles the critical challenge of human-AI complementarity in decision-making. Departing from the traditional focus on algorithmic performance in favor of performance of the human-AI team, and moving past the framing of collaboration as classification to focus on decision-making tasks, we introduce a novel approach to policy learning.  Specifically, we develop a robust solution for human-AI collaboration when outcomes are only observed under assigned actions. We propose a deferral collaboration approach that maximizes decision rewards by exploiting the distinct strengths of humans and AI, strategically allocating instances among them.  Critically, our method is robust to misspecifications in both the human behavior and reward models.  Leveraging the insight that performance gains stem from divergent human and AI behavioral patterns, we demonstrate, using synthetic and real human responses, that our proposed method significantly outperforms independent human and algorithmic decision-making.  Moreover, we show that substantial performance improvements are achievable by routing only a small fraction of instances to human decision-makers, highlighting the potential for efficient and effective human-AI collaboration in complex management settings.
}%

%\KEYWORDS{deterministic inventory theory; infinite linear programming duality;
%  existence of optimal policies; semi-Markov decision process; cyclic schedule}


\maketitle
%%%%%%%%%%%%%%%%%%%%%%%%%%%%%%%%%%%%%%%%%%%%%%%%%%%%%%%%%%%%%%%%%%%%%%

% Samples of sectioning (and labeling) in MNSC
% NOTE: (1) \section and \subsection do NOT end with a period
%       (2) \subsubsection and lower need end punctuation
%       (3) capitalization is as shown (title style).
%
%\section{Introduction}\label{intro} %%1.
%\subsection{Duality and the Classical EOQ Problem}\label{class-EOQ} %% 1.1.
%\subsection{Outline}\label{outline1} %% 1.2.
%\subsubsection{Cyclic Schedules for the General Deterministic SMDP}
%  \label{cyclic-schedules} %% 1.2.1
%\section{Problem Description}\label{problemdescription} %% 2.

% Text of your paper here
\section{Introduction}

\input{intro_r1_revised}

% what is deferral collaboration when AI is not given
% complementarity is not given, but produced by optimizing the AI 
%In many organizational settings, tasks can be strategically allocated to either a human expert or an artificial intelligence (AI) algorithm to leverage the complementary strengths of both. This paradigm, known as deferral collaboration, aims to optimize the performance of the human-AI team by assigning each instance to the partner best suited to handle it \citep{madras2018predict, wilder2020learning}. Research suggests that team performance improves most when an AI system intelligently decides when to defer tasks to its human counterpart \citep{fugener2022cognitive}. The goal is to create a synergistic team whose overall performance surpasses what either the human or the AI could achieve alone.

\section{Background}

In this section, we provide background and related work to help readers better understand the proposed deferral collaboration human-AI system. In \Cref{sec:hai_relatedwork}, we provide an overview of the current research in deferral collaboration, which has thus far focused on classification tasks. In \Cref{sec:prelim}, we provide background on policy learning, which is necessary to understand both the proposed method and the baseline alternatives. We formally introduce notation and present the different policy learning methods, including Direct Method (DM), Inverse-Propensity-Score (IPS) Weighting, and Doubly Robust methods. 

\subsection{Deferral Collaboration}
\label{sec:hai_relatedwork}

Deferral collaboration has considered delegation as an alternative form of human-AI complementarity~\citep{madras2018predict,wilder2020learning,raghu2019algorithmic,de2020regression,fugener2022cognitive}. In such settings, an instance can be assessed by either a human or an AI, the AI is explicitly trained to complement humans, and complementarity is achieved if the instances are routed in such a way that surpasses performance of either entity in isolation. This is the type of complementarity that our work focuses on, and which we refer to as ``deferral collaboration"~\citep{madras2018predict}. 
Prior work has considered the challenge of training such systems in a supervised classification setting \citep{madras2018predict,wilder2020learning,raghu2019algorithmic}, leveraging the prediction uncertainty of the algorithm and the human for each instance. %Notably, the problem of uncertainty quantification for learning from partial feedback remains an open problem. 
To the best of our knowledge, the only work focused on deferral collaboration that considers tasks other than classification is~\citet{de2020regression}, which is focused on regression tasks, but still assumes a traditional supervised learning setting in which each instance has a corresponding label, and this label is observed for all instances in the training set. %The core difference between these works and ours is that they considered contexts where the AI's learning task is a supervised classification task; in this work,  we focus on the problem of learning a personalized decision \emph{policy} that optimally complements a human's decision policy, based on data with bandit feedback. 
Through experimentation with different deferral collaboration mechanisms, \citet{fugener2022cognitive} empirically show that achieving human-AI complementarity is indeed possible, but only if routing is performed by an AI algorithm rather than a human decision maker. 
Our research contributes to this body of work by proposing a deferral collaboration algorithm for policy learning, assuming 
we only have access to the outcomes under assigned treatments in the observational data. 
This enables us to design human-AI collaboration systems for decision-making tasks that involve unknown counterfactuals, such as healthcare \citep{kallus2021minimax}, personalized marketing \citep{elmachtoub2021value}, and customer service \citep{sulek1995impact}, using historical data available in organizational information systems. 
To the best of our knowledge, ours is the first work to tackle this problem.

\subsection{Background on Policy Learning from Observational Data}
\label{sec:prelim}

Policy learning aims to use the observational data to learn a \textit{personalized} decision policy that maps observable individual characteristics to treatments that give favorable rewards. Here, personalized indicates that the treatment is the one that gives the highest reward for a particular instance. One big hurdle to policy learning is the lack of counterfactuals in the observational data, which brings challenges in evaluating and learning a new policy. 

\subsubsection{Problem Statement} 

The observational data consist of $n$ tuples $\mathcal{D} = \{X_i, T_i, Y_i\}_{i=1}^n$, 
where the covariates $X_i\in\mathcal{X}$, the treatment $T_i \in \{0,\cdots, m-1\}$, and a scalar reward $Y_i\in\mathbb{R}$ (higher is better). 
The data is generated by a human behavior policy $\pi_0$ as $T_i \sim \text{Categorical}(\pi_0(T | X_i))$. The behavior policy can be estimated from data. 

Using notations from the Potential Outcome framework in causal inference \citep{rubin2005causal}, we assume $Y_i = Y(X_i,T_i)$, \ie the potential outcome under the observed decision/treatment equals the observed outcome,  known as the SUTVA assumption \citep{rubin1980randomization}. 
The goal is to learn an AI policy $\pi: \mathcal{X} \rightarrow \Delta^m$ where the element in simplex $\Delta^m$ is the probability over the treatment arms, so that the average reward $\mathbb{E}_{T\sim\pi(T|X)}Y(X,T)$ is maximized. 

\subsubsection{AI Policy for Policy Learning from Observational Data} 
Next, we introduce popular AI policy learning methods. Unlike our setting, in which an AI collaborates with a human, in these settings all instances are handled by an AI. We refer to these as AI policies. 

\noindent \textbf{Direct Method (DM):} Direct method fits a machine learning model $f(X,T), f\in\mathcal{F}$ on the observational data $\mathcal{D}$. This can be achieved by minimizing the mean squared loss
$
\min_{f\in\mathcal{F}} \frac{1}{n}\sum_{i=1}^n (Y_i-f(X_i,T_i))^2
$
or other variants \citep{kunzel2019metalearners}. After the reward model is learned, the optimal decisions for future instances will be chosen by 
$
    \arg\max_{T} f(X_i, T)
$ for $X_i$. If $f$ can indeed learn the underlying reward function such that $f(X,T) = E[Y|X,T]$, then direct method will select the optimal decisions for all decision instances. However, due to the selection bias inherently in the observational data and potential model misspecification, direct method is often biased and may have poor performance in practice \citep{kallus2018balanced,elmachtoub2017smart}. 

\noindent \textbf{Inverse Propensity Score Method (IPS):}
Given a candidate policy $\pi$, IPS can evaluate its performance through reweighting the observational data through $\pi_0$. When $\pi_0$ is unknown, we can estimate it by fitting a machine learning model $\hat{\pi}_0$ through cross-fitting~\citep{athey2017efficient}. 
The cross-fitting procedure first splits the data into $K$ folds with equal size. For each fold $k=1, \cdots, K$,  we can fit a machine learning model on the other $K-1$ folds to estimate $\hat{\pi}_{0}^{-k}$, which approximates $\pi_0$ in fold $k$. The cross-fitting is used to avoid overfitting and is essential for efficient estimation in semiparametric models. 
We can then find the optimal policy by optimizing the expected reward of $\pi$:

\begin{align}
\max_{\pi\in\Pi}\sum_{i=1}^N  \frac{\pi(T_i|X_i)}{\hat{\pi}_{0}^{-k(i)}(T_i|X_i)}Y_i, 
\label{eqn:ips}
\end{align}

where $k(i)\in\{1,\cdots,K\}$ is the fold the $i$-th observation belongs to. 
IPS uses the importance sampling trick $\mathbb{E}_{T\sim\pi}Y=\mathbb{E}_{T\sim\pi_0}\frac{\pi(T|X)}{\pi_0(T|X)}Y$ to estimate the reward under target policy $\pi$ and relies on two important assumptions. 
\begin{assumption}[Unconfoundness]\label{ass:unconfoundness}
    $Y(X,T)\perp \!\!\! \perp T | X, \forall\  T \in \{0,\cdots,m-1\}$.
\end{assumption}

\begin{assumption}[Overlap]\label{ass:overlap}
    $\pi_0(T|X)>0, \forall\  T \in \{0,\cdots,m-1\}$.
\end{assumption}

The unconfoundedness assumption means the historical treatments were chosen as a function of $X$ and there are no hidden confounding variables that can influence the treatment and the outcome. 
This assumption is particularly defensible in prescriptive analytics, as suggested in \citet{bertsimas2020predictive}. For example, in certain applications like customer service, the company can access almost any information the human decision makers use to make decisions. There are also recent advances to address the hidden confounding issue \citep{kallus2018confounding,qi2023proximal}, which usually rely on some domain knowledge on the hidden confounding. For example, if we have access to an Instrument Variable (IV), then we can simply incorporate it in the estimation of $\hat{\pi}_0$ \citep{athey2017efficient} and relax the assumption. However, we don't discuss these extensions in detail since it is beyond the scope of this paper. 

The overlap assumption states that humans will have a positive probability to select every treatment. The requirement of this assumption comes from the impossibility result of causal inference \citep{langford2008exploration}. Intuitively, if we never observe the outcome of a certain decision, we cannot make any valid inference about this outcome without strong assumptions. When the overlap assumption is violated, meaning some instances receive deterministic treatments from human decision makers, we can always output the same treatment that human decision makers choose and only do the policy optimization on instances where overlap holds. 
%We discuss our extension in detail in \Cref{app:overlap}. 

Beyond these two assumptions, and similarly to the direct method, when $\hat{\pi}_0$ is misspecified, this objective may be biased and the resulting policy may have bad performance.

\noindent \textbf{Doubly-Robust Method (DR) \citep{dudik2014doubly,athey2017efficient,zhou2023offline}:}
Since DM and IPS both may suffer from model mispecification, it would be ideal if we can use an estimator that is more robust than either of them. The doubly robust (DR) method \citep{dudik2014doubly}, also known as the Augmented IPS \citep{zhou2023offline} method, aims to address that problem. 
DR method also relies on the cross-fitting to estimate both $\pi_0$ and $Y(X,T)$ using machine learning model $\hat{\pi}_0$ and $f$. 
Formally, DR method optimizes the following objective:

\begin{align}
\max_{\pi\in\Pi}\sum_{i=1}^N \Big( \frac{\pi(T_i|X_i)}{\hat{\pi}_{0}^{-k(i)}(T_i|X_i)}(Y_i-f^{-k(i)}(X_i,T_i)) + \sum_T \pi(T|X_i)f^{-k(i)}(X_i,T)\Big). 
\label{eqn:ai_dr}
\end{align}

It can be shown that for the DR objective to be unbiased it is sufficient that \emph{either} $f$ or $\hat{\pi}_0$ is consistent, therefore the DR method offers a ``doubly-robust'' guarantee. We give a formal proof in \Cref{app:proof}. 

In practice, the policy $\pi$ is often chosen to be a linear, tree, or a rule-based policy \citep{athey2017efficient}. Additionally, the policy class $\Pi$ can be constrained to achieve different goals, as discussed in \citet{kitagawa2018should}. For example, given a particular notion of discrimination, one may constrain $\pi$ to not discriminate based on protected features, $\pi$ may consider the budget the requester has, or interpretable policies due to regulation constraints which many institutions require.

\section{Learning Complementary Policies for Human-AI Collaboration \textsc{lcp-hai}}
\label{sec:haimethod}

In this section, we formally define the deferral collaboration problem and introduce the proposed method. In \Cref{sec:hai_ps}, we introduce the objective and framework. In \Cref{sec:hai_drmethod}, we discuss how to develop causal estimators for the deferral collaboration problem, propose a deferral collaboration Doubly-Robust policy learning method, and demonstrate that the proposed estimator is consistent when either the human behavior model or the reward model is consistent. In \Cref{sec:hai_regret}, we present the theoretical regret analysis of the proposed method and show that it enjoys a fast convergence rate.% and is robust to model misspecification. 

\subsection{Problem Statement}
\label{sec:hai_ps}

Compared to the traditional policy learning from observational data, 
\textsc{lcp-hai} is concerned with how to evaluate and learn a human-AI collaborative system which consists of 1) an algorithmic policy 
$\pi : \mathcal{X} \rightarrow \Delta^m$, where the element in simplex $\Delta^m$ is the probability over the treatment arms, along with 2) a routing algorithm $\phi : \mathcal{X} \rightarrow [0,1]$ that chooses whether an instance should be handled by the human or the AI.  We use $\phi(X)$ to denote the probability of routing to humans. Moreover, we assume there is a cost $C(X)$ if a human is chosen to make a decision for task $X$, which reflects the time and resource humans need to make decisions. 

In the context of loan application, for example, when an instance is assigned to AI, the AI decision policy $\pi$ will automatically output a decision to the loan applicant. If a human decision maker is chosen, the human will make an acceptance decision \textit{without AI's help}, which is exactly the same as how the human decision makers historically work before the deferral system is deployed. Therefore, we also assume human decision makers have the same decision performance before and after the deferral system is launched.\footnote{If this assumption is violated and human behavior does change, findings from prior literature suggest that the direction of the change could boost performance further. In particular,~\citet{bondi2022role} shows that when workers are aware that a deferral system is deployed, human decision performance may improve if they are aware that they are selected to solve the tasks because AI thinks they are good at it. In this case, the realized reward of our proposed human-AI system would be even higher than the estimated reward by our objective, as humans would exhibit even higher performance in the instances we assign to them.} 
The final team reward we aim to estimate as a function of $\phi$ and $\pi$ can be written as, 

\begin{align}
\label{eq:exp}
    \theta(\phi, \pi) = \mathbb{E}_{X\sim\mathcal{X},T \sim \pi(T|X)}\phi(X)(Y(X,T) - C(X))  +   (1-\phi(X))Y(X,T). 
\end{align}

The optimal team reward can be written as 
$
    \theta^* = \max_{\phi\in\Phi,\pi\in\Pi} \theta(\phi, \pi)
$. Similarly, the optimal AI routing and decision policy are the ones that maximize $\theta$: $\phi^*, \pi^* = \arg\max_{\phi\in\Phi,\pi\in\Pi}\theta(\phi, \pi)$. The goal in this problem is to learn \emph{both} the routing and decision policies, $\phi,\pi$, that yield the largest reward value possible. This is equivalent to minimize the gap between the performance of the optimal policies and the learned policies, which is often known as regret, formally defined below. 

\begin{definition}[Regret]
    $R(\phi,\pi) = \theta(\phi^*,\pi^*)-\theta(\phi,\pi)$.
\end{definition}

Here, it is worth highlighting one source of complementarity: the humans' historical decision policy $\pi_0$ does not need to be in the policy class $\Pi$. As a consequence, the human and the AI may complement each other by together having a more complex policy class.

\subsection{Develop Human-AI collaboration Methods with Observational Data}
\label{sec:hai_drmethod}

In this section, we introduce the proposed approach to empirically estimate the objective defined in \Cref{eq:exp}, and how to \emph{jointly} optimize the policies $\phi, \pi$.

One possible approach to learning the policies $\phi, \pi$ is to learn them sequentially, which we refer to as Predict, then Collaborate. That is, first learn an AI policy with the goal of achieving the best reward over the entire decision space, and then, given the learned AI decision policy, learn a routing algorithm that allocates instances either to the AI policy or the human based on the learned (human complementing) decision policy. 
Such a strategy serves as an alternative approach we consider, and we introduce it first below.
However, a key innovation of the proposed approach is that we are not merely learning how to allocate tasks between two entities with predefined abilities; rather, the abilities of one of the entities---the AI---are not given but simultaneously being learned specifically to best complement the human's abilities. Thus, we jointly learn the AI policy and the routing algorithm, so that the learning is simultaneously informed by the ability of the AI policy to complement the human as well as the routing policy's ability to discern the superiority of either entity at handing an instance. This offers opportunities for the AI policy to specialize and produce better rewards at decisions the human struggles with than a policy produced to operate autonomously, independently of the human. In this section, we propose two human-AI policy objectives that can be jointly optimized based on Inverse-Propensity-Score-Weighting (IPS) and Doubly Robust (DR) estimators. Below we introduce and discuss each alternative. 

\noindent \textbf{Predict, then Collaborate:} While there has not been human-AI collaboration system proposed for offline policy learning,  a common approach in the literature on deferral collaboration system for classification tasks is to first train an AI model, fix it, and then train a routing algorithm for collaboration \citep{fugener2022cognitive}, which we refer to as the Predict, then Collaborate (PtC) framework. In the context of policy learning, such an approach can be based on the inverse propensity score method. This approach first learns $\pi$ using \Cref{eqn:ips}, and then optimizes the following objective to learn the routing function,

 \begin{align}
\max_{\phi\in\Phi} \hat{\theta}_{\text{P2C}} = \sum_{i=1}^N \phi(X_i)(Y_i - C(X_i))  + \frac{  (1-\phi(X_i))\pi(T_i|X_i)}{\hat{\pi}_{0}^{-k(i)}(T_i|X_i)}Y_i. 
\label{eqn:p2c_ips}
\end{align}

A limitation of this approach that learns the AI decision policy $\pi$ independently of the humans is that it can restrict the potential for complementarity.  Because the AI and humans  may have similar strengths and weaknesses. This overlap in abilities limits the gains from collaboration, as the AI might be focusing on instances where the human already performs well.  Furthermore, independently training an AI to handle all possible instances is often unnecessary and inefficient.  Since the AI will only make decisions on instances routed to it, we can instead tailor its policy to specifically address the gaps in human performance.

To maximize complementarity, we propose jointly learning the human and AI policies. This joint training allows us to explicitly mold the AI policy to excel where humans struggle, even if this means the AI performs suboptimally on instances where humans are already proficient.  The routing algorithm plays a crucial role here.  The opportunities for the AI to complement the human are only meaningful if the routing algorithm can accurately detect this differential performance.  That is, the router must be able to identify instances where the human is likely to struggle and route those instances to the AI. This targeted approach allows the AI to focus its learning on the most impactful areas, leading to a more effective overall system.  By explicitly modeling the interplay between human and AI capabilities, and by leveraging the routing algorithm to direct instances appropriately, we can achieve a more synergistic and performant human-AI team.

\noindent \textbf{Human-AI Collaboration with IPS (HAI-IPS):}
Given a routing algorithm $\phi$ and a target decision policy $\pi$, if we can estimate the future expected human-AI team reward under them, then we can jointly optimize both the routing and decision policy to maximize the future reward. 
One possibility for training such a system is to optimize the empirical estimate of the value function by inverse propensity weighting, using an estimated behavior policy $\hat{\pi}_0$,
 \begin{align}
\max_{\phi\in\Phi,\pi\in\Pi}\hat{\theta}_{\text{IPS}} = \sum_{i=1}^N \phi(X_i)(Y_i - C(X_i))  + \frac{  (1-\phi(X_i))\pi(T_i|X_i)}{\hat{\pi}_{0}^{-k(i)}(T_i|X_i)}Y_i. 
\label{eqn:hai_ips}
\end{align}

The first term in Equation~\ref{eqn:hai_ips} is the expected human reward when humans are chosen for making decisions, i.e. when $\phi(X_i)=1$, and the second term is the expected reward for policy $\pi$. 
However, HAI-IPS shares the same issue as IPS when $\pi_0$ can only be learned through a slow rate \citep{d2021overlap,zhou2023offline} or $\hat{\pi}_0$ is misspecified. 

\noindent \textbf{Human-AI Collaboration with DR (HAI-DR):} 
We propose a doubly-robust objective to augment HAI-IPS with the direct method, which leads to the following optimization problem: $\max_{\phi\in\Phi,\pi\in\Pi} \hat{\theta}_{\text{DR}} = $

{\small
 \begin{align}
\sum_{i=1}^N \phi(X_i)(Y_i - C(X_i))  + (1-\phi(X_i)) \Big(\frac{\pi(T_i|X_i)}{\hat{\pi}_{0}^{-k(i)}(T_i|X_i)}(Y_i-f^{-k(i)}(X_i,T_i)) + \sum_T \pi(T|X_i)f^{-k(i)}(X_,T) \Big). 
\label{eqn:hai_dr}
\end{align}
}

The first term estimates the human performance under a future routing policy $\phi$, and the second term estimates the performance of a future decision policy $\pi$ under a future routing policy $\phi$. Since the decisions made by the future decision policy $\pi$ can be different from the decisions from the previous human policy $\pi_0$, we can use policy evaluation methods \citep{dudik2014doubly,athey2017efficient,zhou2023offline} to estimate the future performance of $\pi$. 

The newly proposed doubly-robust (DR) objective benefits from the doubly-robustness property, which ensures that our estimator, denoted as $\hat{\theta}_{\text{DR}}$
 , remains consistent under weaker assumptions than $\hat{\theta}_{\text{IPS}}$. Specifically, $\hat{\theta}_{\text{DR}}$ is guaranteed to provide a consistent estimate of the reward in the human-AI system as long as \emph{either} the propensity score model $\pi_0(T|X)$ (which estimates the probability of taking a particular action given the context) or the reward model $f(X,T)$ (which estimates the expected reward given the action and context) is correctly specified. Crucially, we do not need to know which of the two is misspecified. 
 This robustness to model misspecification makes the doubly-robust approach particularly advantageous in practical scenarios where only one of the models (propensity score or reward) may be accurately estimated, but not necessarily both. As a result, the DR estimator can mitigate the biases that arise when either the propensity score or reward model is misspecified, ensuring reliable performance even in the presence of imperfect modeling assumptions. The doubly-robust guarantee is formally stated in \Cref{prop:dr} and the proof is included in \Cref{app:proof}.

\begin{proposition}[Doubly-Robustness]
\label{prop:dr}
    Under \Cref{ass:unconfoundness}, \Cref{ass:overlap}, and SUTVA, for a given routing and decision policy $\phi,\pi$, 
    $\hat{\theta}_{\text{DR}}$ is a consistent estimator of $\theta$ if $f$ is a consistent estimator of $Y(X,T)$ or $\hat{\pi}_0$ is a consistent estimator of $\pi_0$.
\end{proposition}

\noindent \textbf{Optimization of the Human-AI System:} 
Having established the optimization objective, we now discuss its practical implementation for various policy classes $\Phi,\Pi$. For finite policy classes, exhaustive search across all policy combinations offers a direct route to maximizing the estimated reward in Equation \eqref{eqn:hai_dr}. However, this approach becomes computationally intractable for complex policy spaces.  Therefore, we focus on differentiable policy classes, such as logistic policies \citep{kallus2021minimax}, which enable efficient optimization via gradient descent.  This allows us to simultaneously optimize both the AI decision policy and the routing policy.  Furthermore, we note that for tree-based policy classes, the optimization problem can be formulated as a mixed integer program, as demonstrated in \citet{zhou2023offline}.

\subsection{Regret Guarantees}
\label{sec:hai_regret}

In this section, we establish regret guarantees for the learned routing and decision policies. 
We show that the convergence rate of the regret, i.e. the gap between the reward of the optimized policy and the reward of the optimal policy, is ${O}(\frac{1}{\sqrt{n}})$. This indicates that the reward of the policy learned from the doubly robust estimator will converge to the optimal reward at a fast speed when the size of the data samples increases. All proofs for this section are included in \Cref{app:proof}. 

To establish the regret guarantee, 
we assume that the reward model and the human behavior model follow the consistency assumption. This assumption is commonly used in the causal machine learning literature \citep{zhou2023offline} and is a generalization of the assumptions in \citet{chernozhukov2017double}. Note that this assumption differs from the assumption needed to prove double-robustness in~\Cref{sec:hai_drmethod}, which only requires either of these to be consistent. 

\begin{assumption}[Consistency] \label{ass:consistency}
    $f(X,T), \hat{\pi}_0(X)$ estimated on $n$ samples follow the error bounds
    \begin{align}
        \mathbb{E}[(f(X,t)-Y(X,t))^2]\mathbb{E}[(\hat{\pi}_0(t|X)-\pi_0(t|X))^2]=\frac{o(1)}{n}, \forall t \in 0, \cdots, m-1
    \end{align}
\end{assumption}

We define the true reward gap and the estimated reward gap between two different deferral collaboration systems as $\Delta(\{\pi_1,\phi_1\},\{\pi_2,\phi_2\})$ and $\hat{\Delta}(\{\pi_1,\phi_1\},\{\pi_2,\phi_2\})$, respectively.

\begin{definition}
    $\Delta(\{\pi_1,\phi_1\},\{\pi_2,\phi_2\}) = \theta(\{\pi_1,\phi_1\})-\theta(\{\pi_2,\phi_2\})$
\end{definition}

\begin{definition}
    $\hat{\Delta}(\{\pi_1,\phi_1\},\{\pi_2,\phi_2\}) = \hat{\theta}(\{\pi_1,\phi_1\})-\hat{\theta}(\{\pi_2,\phi_2\})$
\end{definition}

Next, we define $\tilde{\theta}$ as the empirical estimator with the true nuisance functions, i.e. the reward model and the human behavior model,
\begin{align}
\tilde{\theta}(\{\pi, \phi\}) = \frac{1}{N}\sum_{i=1}^N \phi(X_i)(Y_i - C(X_i))  + (1-\phi(X_i)) \Big(\frac{\pi(T_i|X_i)}{{\pi}_{0}(T_i|X_i)}(Y_i-f(X_i,T_i)) + \sum_T \pi(T|X_i)f(X,T) \Big). \nonumber  
\label{eqn:tilde_hai_dr}
\end{align}

\begin{definition}
    $\tilde{\Delta}(\{\pi_1,\phi_1\},\{\pi_2,\phi_2\}) = \tilde{\theta}(\{\pi_1,\phi_1\})-\tilde{\theta}(\{\pi_2,\phi_2\})$
\end{definition}

Let $\hat{\pi},\hat{\phi} = \arg\max_{\pi\in\Pi,\phi\in\Phi}\hat{\theta}_{\text{DR}}$, which is the deferral collaboration system (AI policy and router) optimized from the empirical doubly robust estimator, as defined in~\Cref{eqn:hai_dr}. We want the regret to converge to 0 at a fast rate (e.g., $O(\frac{1}{\sqrt{n}})$), which would mean the proposed algorithm has a good performance. 

In order to characterize the regret bound, we will first present standard complexity definitions for various policy classes \citep{zhou2023offline}. %, which are crucial for characterizing the regret bound in our analysis. 
These measures are mathematical tools designed to quantify the richness or expressiveness of a policy class. Intuitively, they capture how flexible a policy class is in fitting different datasets or decision environments. The greater the complexity of a policy class, the higher its capacity to adapt to diverse scenarios—but this also increases the risk of overfitting and impacts the theoretical bounds on regret. By incorporating these complexity measures, we can rigorously evaluate the trade-off between model flexibility and performance guarantees, providing a structured way to analyze the regret bound. 

\begin{definition} 
    Let $Z_i$ be the iid Rademacher variables ($P(Z_i=1)=P(Z_i=-1)=0.5$)\citep{zhou2023offline}. The empirical Rademacher complexity is defined as 
    \begin{align}
        \mathcal{R}_n(\Pi^D;\{X_i,\Gamma_i\}_{i=1}^n) = 
        \mathbb{E}[\sup_{\pi_1,\pi_2 \in\Pi} \frac{1}{n}|\sum_{i=1}^n Z_i\langle\Gamma_i,\pi_1(X_i)-\pi_2(X_i)\rangle|\{X_i,\Gamma_i\}_{i=1}^n]
    \end{align}
    where the expectation is taken with respect to $\{Z_i\}_{i=1}^n$.
    The Rademacher complexity $\mathcal{R}_n(\Pi^D)$ is taken with respect to the samples $\mathcal{R}_n(\Pi^D) = \mathbb{E}[\mathcal{R}_n(\Pi^D;\{X_i,\Gamma_i\}_{i=1}^n)]$.
\end{definition}

\begin{definition}
    For feature domain $\mathcal{X}$, the policy class $\Pi$, a set of $n$ points $\{X_1, \cdots, X_n\} \subset \mathcal{X}$, define 
\begin{enumerate}
    \item Hamming Distance between two policies $\pi_1$ and $\pi_2$ as $H(\pi_1,\pi_2)=\frac{1}{n}\sum_{i=1}^n \mathbb{I}[\pi_1(X_i)\neq \pi_2(X_i)]$. 
    \item $\epsilon$-Hamming covering number of the set $N_{HS}(\epsilon,\Pi,\{X_1, \cdots, X_n\})$ as the smallest number of policies in $\Pi$, such that $\forall \pi\in\Pi, \exists \pi_i, H(\pi,\pi_i)\leq\epsilon$. $\epsilon$-Hamming covering number of the policy class $\Pi$ is $N_H(\epsilon,\pi)=\sup\{N_{HS}(\epsilon,\Pi,\{X_1, \cdots, X_n\})|m\geq 1, X_1, \cdots, X_m\in\mathcal{X}\}$. 
    \item The entropy integral is $\textit{k}(\Pi) = \int_0^1 \sqrt{\log N_H(\epsilon^2,\pi)}d\epsilon$. 
\end{enumerate}
\end{definition}

The entropy integral of the policy class $\Pi$ as $\textit{k}(\Pi)$ is a complexity measure of the policy class $\Pi$ that can be used to quantify the Rademacher complexity. This is a variant of the classical entropy integral \citep{dudley1967sizes} and we give a detailed discussion of its definition in \Cref{app:proof}. Next, we make an assumption about the covering number of the policy class, which is a standard assumption that holds for common policy classes \citep{zhou2023offline}. 

\begin{assumption}\label{ass:growthrate}
$\forall 0<\epsilon<1, N_H(\epsilon, \Pi)\leq C\exp(D(\frac{1}{\epsilon})^w)$ for constants $C,D>0, 0<w<0.5$. 
\end{assumption}

In addition, we use a simplified notation for $\tilde{\theta}$ that $\tilde{\theta}(\{\pi, \phi\}) = \frac{1}{N}\sum_{i=1}^N\phi(X_i)(Y_i - C(X_i))  + (1-\phi(X_i)) \langle \Gamma_i, \pi(X_i)\rangle$, where $\Gamma_i = \frac{Y_i(T_i)-f(X_i,T_i)}{\pi_0(X_i,T_i)}\mathbf{1}[T=T_i] + \left(\begin{smallmatrix}f(X_i,T_1) \\ \cdots\\f(X_i,T_m)\end{smallmatrix}\right)$. 

With the key tools and definitions in place, we now turn to bounding the Rademacher Complexity of the policy class, a critical step in the regret analysis. The Rademacher Complexity is a specific complexity measure of the policy class that quantifies its ability to fit random noise. A tighter bound on this complexity helps us control the generalization error, ensuring that the learned policy performs well not just on the observed data but also in unseen scenarios. By deriving an explicit bound for the Rademacher Complexity, we lay the foundation for rigorously analyzing the regret. 

\begin{lemma}[Rademacher Complexity]
\label{lemma:rademacher}
Under \Cref{ass:growthrate} with iid bounded $\{\Gamma_i\}_{i=1}^n$, we have 
    \begin{align}
        \mathcal{R}_n(\Pi^D) \leq 27.2(\textit{k}(\Pi)+8)\sqrt{\frac{\sup_{\pi_1,\pi_2\in\Pi }\mathbb{E}[\langle\Gamma_i,\pi_1(X_i)-\pi_2(X_i)\rangle^2]}{n}}
        + o(\frac{1}{\sqrt{n}})
    \end{align}
\end{lemma}

\Cref{lemma:rademacher} is from \citet{zhou2023offline} to measure the complexity of different policy classes. This bound only depends on the magnitude of the difference induced by two policies. Given the overlap condition and a bounded reward, the difference in magnitude between two policies can be upper bounded by a problem-dependent constant.

Given the Rademacher Complexity, we now proceed to derive the regret bound. The high level proof goes as follows. The regret $R(\hat{\pi},\hat{\phi})$ can be decomposed into two key terms: $\sup|\Delta-\tilde{\Delta}|$ and $\sup|\hat{\Delta}-\tilde{\Delta}|$. In the subsequent lemmas, we provide separate bounds for each term and demonstrate that both converge at a rate of $O(\frac{1}{\sqrt{n}})$. Combining these results yields the final regret bound, capturing the asymptotic behavior of the policy's performance. 

\begin{lemma}\label{lemma:tildedelta}
Under Assumptions of \Cref{lemma:rademacher}, we have 
    \begin{align}
&        \sup_{\substack{\pi_1,\pi_2 \in\Pi\\ \phi_1,\phi_2\in\Phi}}
|\Delta(\{\pi_1,\phi_1\},\{\pi_2,\phi_2\})-\tilde{\Delta}(\{\pi_1,\phi_1\},\{\pi_2,\phi_2\})| \nonumber \\
& \leq 71.6 \sqrt{2}(\textit{k}(\Pi)+8)\sqrt{\frac{\sup_{\pi_1,\pi_2\in\Pi }\mathbb{E}[\langle\Gamma_i,\pi_1(X_i)-\pi_2(X_i)\rangle^2]}{n}} \nonumber \\
& \quad + 54.4\sqrt{2}(\textit{k}(\Phi)+8)\sqrt{\frac{\sup_{\phi_1,\phi_2\in\Phi }\mathbb{E}[\langle Y_i,\phi_1(X_i)-\phi_2(X_i)\rangle^2]}{n}} + o(\frac{1}{\sqrt{n}})
    \end{align}
\end{lemma}

\begin{lemma}
\label{lemma_hat_delta_rate}
\begin{align}
        \sup_{\substack{\pi_1,\pi_2 \in\Pi\\ \phi_1,\phi_2\in\Phi}}
|\hat{\Delta}(\{\pi_1,\phi_1\},\{\pi_2,\phi_2\})-\tilde{\Delta}(\{\pi_1,\phi_1\},\{\pi_2,\phi_2\})| = o_p(\frac{1}{\sqrt{n}})
\end{align}
\end{lemma}

\begin{theorem}[Regret Convergence]\label{thm:main} Under Assumptions 1, 2, 3 and the Assumptions in \Cref{lemma:rademacher}, we have 
\begin{align}
    R(\hat{\pi},\hat{\phi}) = O_p((\textit{k}(\Pi)+\textit{k}(\Phi))\sqrt{\frac{V^*}{n}})
\end{align}
\end{theorem}

This theoretical result suggests that for most common policy classes such as the logistic policy or tree policies with finite depth, the regret of the proposed deferral collaboration system compared to the optimal system decreases following $O_p(\frac{1}{\sqrt{n}})$ with respect to $n$. This suggests that the proposed doubly robust estimator can find the best human-AI system in an optimal rate while all nuisance components are estimated at $O_p(\frac{1}{n^{0.25}})$ in root-mean-squared error \citep{athey2017efficient,zhou2023offline}.

\section{Best AI may not be the best human teammate}
\label{sec:best_vs_complementary}

In a deferral collaboration system, the AI that achieves the highest average reward in the policy class $\Pi$ may not necessarily be the one that best enables the human-AI team to achieve the optimal team performance. The idea that the best-performing individual is not the best teammate has been discussed in the context of human teams and cognitive diversity, where it has been noted that organizations can improve their performance by building teams with members that complement one another~\citep{page2019diversity}. The deferral collaboration literature has shown this to also be the case when a team made up of a human and an AI engage in a classification task \citep{wilder2020learning}. In this section, we provide a simple example to give intuition for why this is also the case in the context of policy learning.   %Intuitively, an AI with slightly lower independent performance could form a better human-AI system if it excels in decision instances where the human performs poorly. In contrast, an AI that performs similarly to the human across all tasks, whether good or bad, may offer less complementary value, despite its higher standalone performance.

\begin{figure}
    \centering
    \begin{subfigure}[b]{0.3\textwidth}
        \includegraphics[width=\textwidth]{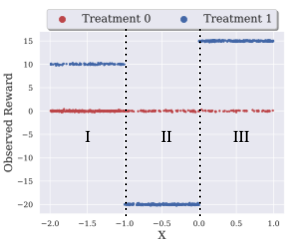}
        \caption{Potential Outcomes}
        \label{fig:syndata}
    \end{subfigure}
    \begin{subfigure}[b]{0.33\textwidth}
        \includegraphics[width=\textwidth]{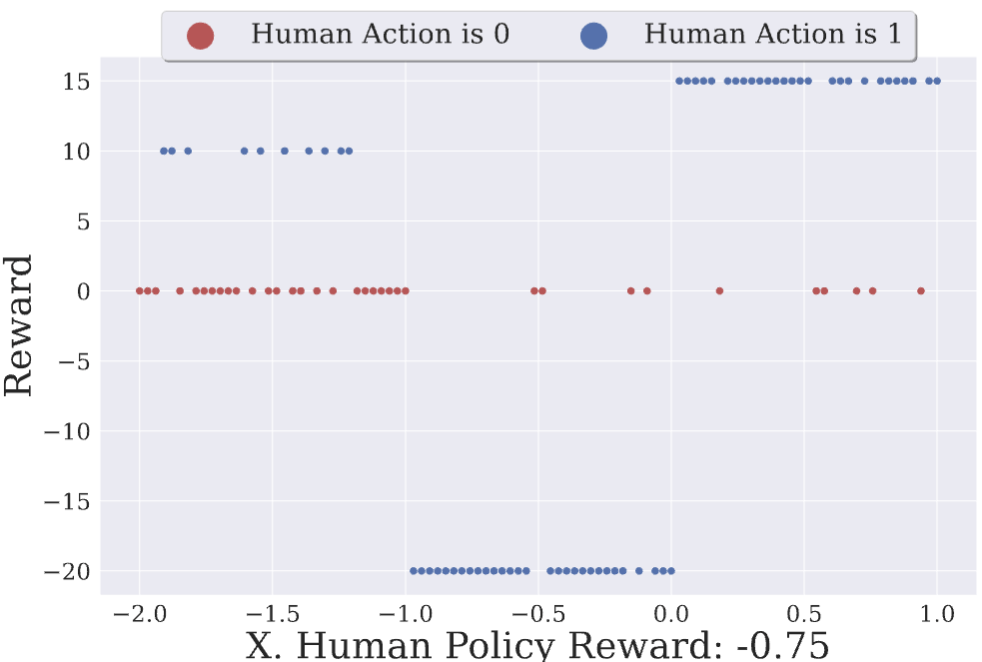}
        \caption{Human Policy}
        \label{fig:humanpolicy}
    \end{subfigure}
    \begin{subfigure}[b]{0.33\textwidth}
        \includegraphics[width=\textwidth]{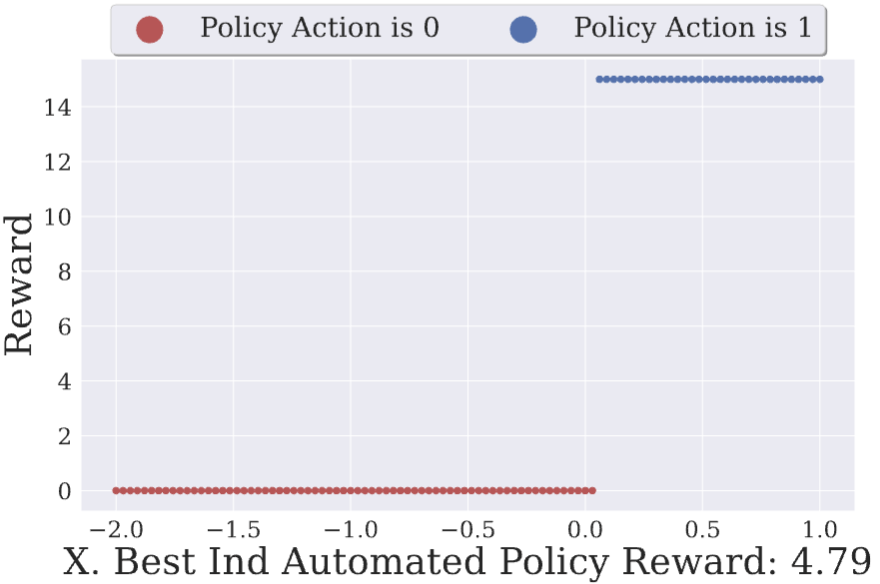}
        \caption{AI Policy}
        \label{fig:aipolicy}
    \end{subfigure} \\
    \begin{subfigure}[b]{0.35\textwidth}
        \includegraphics[width=\textwidth]{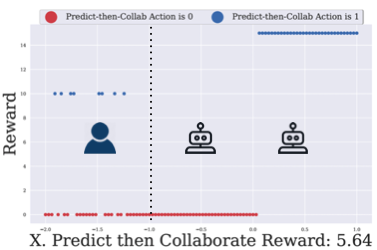}
        \caption{Predict-then-Collaborate}
        \label{fig:ptc}
    \end{subfigure}
    \begin{subfigure}[b]{0.35\textwidth}
        \includegraphics[width=\textwidth]{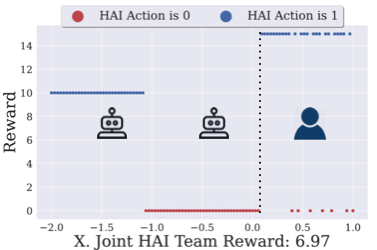}
        \caption{Joint HAI}
        \label{fig:hai}
    \end{subfigure}
    \caption{Illustration of Different Policies on the Synthetic Data. (a): The potential outcomes of the data-generating process; (b): Observed reward under the human policy; (c): Observed reward under the best independent AI policy; (d): Observed reward under the Predict-then-Collaborate system; (e): Observed reward under the joint Human-AI system.}
    \label{fig:synall}
\end{figure}

For the illustrative example, we show both the theoretical optimal solutions under each method and the empirical solutions optimized on a corresponding observational data using baselines and our proposed human-AI teams. The theoretical analysis provides insights on when the complementarity arises and is maximized. The empirical analysis complements the theoretical analysis by visualizing this example in \Cref{fig:synall} and showing the proposed methods can achieve the theoretical optimal performance given only the observational data. 

Given  \Cref{ass:unconfoundness}, \Cref{ass:overlap}, and \Cref{ass:consistency}, it is known that the interventional mean $E[Y|X,T]$ is identifiable from the observational data with enough data samples, following the proof in \Cref{prop:dr}. Therefore, for the theoretical analysis in this section, we assume that we have the true conditional expected reward information, which allows us to simplify the discussion without the approximation error term. 

For visualization purposes, in our example, the decision instances' features have a one-dimensional embedding $X\in[-2,1]$. 
The potential outcomes follow the following distribution:

\begin{align}
    Y(X,T=1) = \begin{cases}
        \delta_1, & -2\leq X\leq -1, \quad \text{Region I} \\
        -\delta_2, & -1\leq X\leq 0, \quad \text{Region II} \\
        \delta_3, & 0\leq X\leq 1, \quad \text{Region III} \\
    \end{cases}
\end{align}

and $Y(X,T=0)\equiv 0$, where $\delta_1,\delta_2,\delta_3>0$. $X$ follows a uniform distribution on $[-2,1]$. While the analysis is general for a wide range of $\delta_1, \delta_2,$ and $\delta_3$, we provide a specific realization of this example in \Cref{fig:syndata} with $\delta_1 = 10, \delta_2=20, \delta_3 = 15$. In the empirical visualization, the potential outcomes have an additional noise $\epsilon \sim \mathcal{N}(0,0.1)$. 

The human decision maker will select the optimal decisions with the following probability:
\begin{align}
    \pi_0(T=1|X) = \begin{cases}
        \rho_1, & -2\leq X\leq -1, \quad \text{Region I} \\
        \rho_2, & -1\leq X\leq 0, \quad \text{Region II} \\
        \rho_3, & 0\leq X\leq 1, \quad \text{Region III} \\
    \end{cases}
\end{align}

It is easy to see that the human reward is $(\rho_1\delta_1-\rho_2\delta_2+\rho_3\delta_3)/3$. In the empirical realization of human actions in \Cref{fig:humanpolicy}, we set $\rho_1=0.3,\rho_2=\rho_3=0.8$. 
For simplicity, we assume $C(X)\equiv 0$ and the requester adopts a linear policy class for both the router and the decision policy (similar data generating processes can be derived for nonzero costs). 
In this case, each linear policy allows to use one point $X_0$ to switch decisions. 

Assume $\delta_2>\delta_1$, $\delta_3>\delta_1$, and $(1-\rho_1)\delta_1>(1-\rho_3)\delta_3$. In this setting, the AI policy that best complements the human decision maker to maximize joint performance is not the AI policy that yields the best independent performance. The best AI policy working alone would choose $T=1$ when $X>0$ and $T=0$ otherwise, as shown in \Cref{fig:aipolicy}, which leads to a reward of $\delta_3$. 
The human-AI team performance when working with this AI is $\delta_3+\rho_1\delta_1$, since the optimal router will give $X>-1$ to AI and all other instances to the human, as shown in \Cref{fig:ptc}. Meanwhile, a complementary AI would choose $T=1$ when $X<-1$ and $T=0$ otherwise, which leads to a reward of $\delta_1$. The optimal router would only queries the human when $X>0$, so the human-AI team has a reward $\delta_1+\rho_3\delta_3$. This is shown in \Cref{fig:hai}. From the assumption $(1-\rho_1)\delta_1>(1-\rho_3)\delta_3$, we can see that the resulting human-AI team considering human behavior is better than the human working with the best independent AI. We also note that a wide range of $\delta_1,\rho_1,\delta_2,\rho_2, \delta_3, \rho_3$ can satisfy this condition. 

A detailed illustration of how the best human-AI team is selected is shown in \Cref{tab:best_vs_comp}. 
For simplicity in presentation, we multiply the reward by a factor of 3. 
In this setting, there are 5 possible linear policies and we show three of them in the table. Intuitively, we compare the human-AI team performance under every potential AI policy and pick the best one. Among all potential policies, the Potential AI 1 achieves the best human-AI team performance while having a worse individual performance compared to the best independent AI. 

\begin{table}[]
    \centering
    \begin{tabular}{c|ccccc}
        $P(T=1|X)$ &  Human & Best Ind AI  & Potential AI 1 & Potential AI 2 & ...\\
        I & $\rho_1$ & 0 & 1 &1 \\
        II & $\rho_2$ & 0 & 0&1\\
        III & $\rho_3$ & 1& 0 &0\\
        reward & $\rho_1\delta_1-\rho_2\delta_2+\rho_3\delta_3$& \boldmath${\delta_3}$ & $\delta_1$&$\delta_1-\delta_2$\\
       w/ Opt Routing &   & + Human  & + Human & + Human & ...\\
        I & - & $\rho_1$ & 1 & 1\\
        II & - & 0 & 0 & $\rho_2$\\
        III & - & 1 & $\rho_3$ & $\rho_3$\\
        reward & - & $\delta_3+\rho_1\delta_1$& \boldmath$ \delta_1+\rho_3\delta_3$&$\delta_1-\rho_2\delta_2+\rho_3\delta_3$\\
    \end{tabular}
    \caption{Human-AI Collaboration under the Stylized Model with Full Information.}
    \label{tab:best_vs_comp}
\end{table}

The corresponding empirical realizations for different methods are shown in \Cref{fig:synall}. To reiterate, 
\Cref{fig:syndata} shows the data distributions of the potential outcomes under Region I, II, and III. 
In \Cref{fig:humanpolicy} and \Cref{fig:aipolicy}, we visualize the observed reward under the human policy and the best independent AI policy, respectively. The AI policy has a reward of 4.79, which is significantly better than the human's reward of -0.75. Under the best independent AI policy, the routing algorithm will assign the tasks in Region I to the human, yielding a team performance of 5.64, as shown in \Cref{fig:ptc}. However, the Joint HAI Team will assign the tasks in Region III to the human and adopt an AI policy that is slightly worse than the best independent AI policy when assessed individually; this system yields the highest team reward of 6.97. Through this toy example, we can see that the best independent AI may not the best human teammate. Notably, compared to the Predict-then-Collaborate framework, a jointly optimized human-AI system can help find the complementary AI policy that gives the best team reward, emphasizing the value of jointly learning the AI policy and the router. 

\noindent\textbf{What instances are routed to humans?} If the optimization is perfect, for every potential AI policy, our algorithm will compare the conditional expected reward between the human and the AI policy in a data-driven way. For example, since the human has a better decision making performance than the best independent AI ($\rho_1\delta_1>0$) in region I. Practically, this implies that humans should have a better decision performance in such instances compared to the given AI policy. This could imply humans have a different and more powerful decision policy on such decision instances. 

\section{Experiments}

In this section, we empirically evaluate the efficacy of the proposed method. First, we use the synthetic data introduced in \Cref{sec:best_vs_complementary} to demonstrate the proposed method's double robustness property compared to alternative methods; we present these results in \Cref{sec:exp_syndata}. In \Cref{sec:exp_ihdp}, we use a real-world example commonly used in the causal inference literature, which considers the problem of assigning home visit to infants in order to improve future cognitive test scores. Consistent with prior literature, these experiments simulate the past decisions in the observational data. In \Cref{sec:exp_webshop}, we demonstrate the benefit of our proposed method using real human behaviors in an online shopping assistant task. Additionally, \Cref{app:addexp} presents evaluations using a common alternative used in policy learning \citep{zhan2024policy,swaminathan2015counterfactual,dudik2014doubly} that adapts multiclassification datasets to policy learning problems by treating each class as an arm and the one-hot encoding of the associated label as the counterfactual outcome of different arms. 

\subsection{Baselines}

We compare our methods HAI-IPS (\Cref{eqn:hai_ips}) and HAI-DR (\Cref{eqn:hai_dr}) with the following baselines:

\begin{itemize}
    \item Human: Every decision is made by a human decision maker. 
    \item Direct Method (DM-AI): A machine learning model $f(X,T)$ is fitted to estimate $E[Y|X,T]$ by minimizing the mean squared error on the observational data. All instances are handled by the AI policy.
    \item Doubly-Robust (DR-AI): The doubly robust estimator selects the best AI policy through optimizing \Cref{eqn:ai_dr}. All instances are handled by the AI policy.
    \item Predict-then-Collaborate-DR: First, an AI policy is optimized by minimizing the DR objective. This AI policy is fixed, and then the routing model is optimized to determine optimal allocation between this AI policy and humans.
\end{itemize}

In \Cref{sec:exp_syndata}, in addition, we also compare with the following baselines, which allow us to assess the benefits of the proposed doubly robust approach compared to alternative architectures:

\begin{itemize}
    \item AI simulated Human + AI Policy (AI-AI): Considers the solution returned by HAI-DR and replaces the human with the actions predicted by the human behavior model, yielding AI-AI collaboration. 
    \item DM (Complex Policy): An AI-only policy, where the policy is allowed to follow a more complex policy, %that is decided by the direct method and 
    not contained in the policy class $\Pi$. In our experiments, we use a gradient boosting tree to fit $f(X,T)$ and select actions based on $\arg\max_T f(X,T)$ for $X$. 
\end{itemize}

As we shall see in \Cref{sec:exp_syndata}, the AI-AI baseline has a poor performance when the human behavior model is misspecified, and the DM (Complex Policy) baseline has a poor performance when the direct method is misspecified, even if the policy class is allowed to be much more complex. 

We note that there are many works that have proposed variations to improve the direct method \citep{johansson2016learning,shalit2017estimating,gao2021enhancing} and the IPS method \citep{swaminathan2015self,kallus2018balanced}. The proposed deferral collaboration framework can easily incorporate and benefit from these extensions (e.g., by replacing $f(X,T)$ to another direct method estimator or by changing the IPS estimator to the self-normalized estimator). Therefore, we only compare with the most widely used DM and IPS estimators throughout the experiments.

\subsection{Double Robustness}
\label{sec:exp_syndata}

Recall the synthetic data introduced in \Cref{sec:best_vs_complementary}. We set $\rho_1=0.3,\rho_2=\rho_3=0.8$, $\delta_1 = 10, \delta_2=20, \delta_3 = 15$. As in~\Cref{sec:best_vs_complementary}, we assume the potential outcome has noise $\epsilon\sim \mathcal{N}(0,0.1)$. The resulting data distribution is shown in \Cref{fig:syndata}. We assume all algorithms only have access to the observational data, which is produced by historical human decisions. Similar to our discussion in \Cref{sec:best_vs_complementary}, the policy class is chosen to be a linear policy class $P_{a,b}(T=1)=\sigma(ax + b)$, where $\sigma(\cdot)$ is the sigmoid function and $a,b$ are the parameters to be optimized. 

We report the rewards of different methods in \Cref{tab:synreward}. The rewards are reported under three conditions, specifically designed to exhibit the doubly-robust properties of the proposed method. It is worth noting that, while in this case we can know whether a model is correctly specified because we have knowledge of the models we used to generate the synthetic data, this knowledge is typically unavailable in practice. The first condition we consider is (1) DM Correct: under this condition, we use a gradient boost tree (can be thought of as a non-parametric learner) as the reward model. For this data, this model specification can capture the true distribution and is therefore correctly specified. Meanwhile, we fit the propensity score model using a logistic regression model, which will wrongly estimate the true human behaviors. The second condition is (2) PS-Correct: we use the logistic regression model as the reward model (misspecified) and the gradient boosting tree as the propensity score model. Finally, we consider (3) Worst Case: we consider the worst case scenario, which is the worst of the first two cases, and report it. Note that this last scenario captures the fact that, in practice, we would not know which of the two may be misspecified. 

As shown in the results in~\Cref{tab:synreward}, when the reward model is correctly specified, HAI-DR achieves the best performance and is better than all other alternatives. In particular, it is significantly  better than HAI-IPS (p-value under paired t-test: 0.0003) due to the misspecification in the human behavior model. When the human behavior model is correct, HAI-IPS and HAI-DR yielding statistically comparable performance (p-value under paired t-test: 0.38) and the best performance compared to other methods, and direct method has a significant performance drop.

\begin{table}[!ht]
    \centering
    \begin{tabular}{l|l|l|l}
        ~ & Reward Model Correct & Human Behavior Model Correct & Worst Case \\ \toprule 
        DM & $4.86 (0.11)$ & $4.07 (0.47)$ & $4.07$ \\ 
        Human & $-0.38 (0.06)$ & $-0.36 (0.05)$ & $-0.38$ \\ 
        DR-AI & $4.86 (0.11)$ & $4.82 (0.06)$ & $4.82$ \\ 
        HAI-IPS & $5.41^* (0.36)$ & $6.49^*(0.94)$ & $5.41$ \\
        Predict-then-Collaborate-DR & $5.18 (0.45)$ & $5.13 (0.46)$ & $5.13$ \\ 
        HAI-DR & $7.01^* (0.07)$ & $6.91^* (0.1)$ & $\textbf{6.91}$ \\ \midrule 
        AI-AI & $5.80^* (0.02)$ & $6.92^*(0.11)$ & $5.8$ \\
        DM-(Complex Policy) & $8.29^* (0.03)$ & $4.01 (0.49)$ & $4.01$ \\ \bottomrule
    \end{tabular}
    \caption{Reward of Different Methods. $*$ indicates the reward is significantly better than both the Human's reward and the DR-AI's reward at the $\alpha=0.05$ significance level.}
    \label{tab:synreward}
\end{table}

\subsection{Infant Health and Development Program (IHDP)}
\label{sec:exp_ihdp}

How to benchmark policy learning and causal inference methods is a challenging open problem since counterfactuals are never in observational datasets \citep{parikh2022validating}. Moreover, note that while randomized control trials pose an avenue for gaining knowledge about counterfactuals, in such datasets we do not observe natural data generating processes that are of relevance to some methods, like ours. Specifically, our method aims to learn from historical human decisions, which are unavailable in randomized control trials as the human decision is replaced by the randomization mechanism. Therefore, we adopt the common practice in the causal inference literature of using semi-synthetic datasets to validate the proposed methods. Due to the lack of counterfactuals, researchers in causal inference use real-world data and simulate a response curves given the covariates to validate causal inference methods \citep{johansson2016learning,shalit2017estimating,louizos2017causal,yoon2018ganite,zhang2022exploring}. This allows researchers to systematically evaluate methods against the ground truth, as the true relationships between actions and rewards are fully controlled in the simulation. This strategy has also been adopted by policy learning researchers \citep{kallus2021minimax,zhou2023offline,zhan2024policy}. 

Infant Health and Development Program (IHDP) is a randomized experiment from 1985 to 1988 which studied the effect of home visits on cognitive test scores for infants. The dataset is first used for benchmarking causal machine learning algorithms using synthetic response curves in \citet{hill2011bayesian} and widely adopted in causal machine learning papers \citep{johansson2016learning,shalit2017estimating,louizos2017causal}, where rewards are generated using original covariates. The rewards refer to the cognitive test scores, and the decision in whether to assign a home visit to an infant. Here, we use the dataset for evaluating different policy learning methods, and considering whether human-AI collaboration may outperform either alternative alone.

The dataset contains 747 subjects and 25 variables, where the variables represent different aspects of children and their mothers. For the synthetic response, we assume $Y_0 = \beta^T X$, where every element of $\beta$ is sampled from $[0,0.1,0.2,0.3,0.4]$ with probability $[0.6,0.1,0.1,0.1,0.1]$, following \citet{hill2011bayesian}. 
We assume the treatment effect $Y_1-Y_0=10(X_1^2-3)$, where $X_1$ is the second element of $X$. For the human policy, we assume humans will adopt the optimal action with probability 0.8 when $X_1>0$, and this probability will decrease to 0.4 when $X_1\leq 0$.

The results of the different methods are shown in \Cref{tab:ihdp}. HAI-DR has the best reward among all methods. While the human policy has a much lower reward the best independent AI, the joint Human-AI team has a much better performance than the independent AI. To better understand how often instances are routed to the human and the AI, we show the deferral statistics in \Cref{tab:deferral_ihdp}. Interestingly, the human-AI system achieves the significant performance improvement with only 3\% of the instances being routed to the human. When the instances are routed to the human, the average human reward (1.94) is significantly better than the AI's reward (0.12). When the tasks are assigned to the AI, the AI's reward (3.16) is better than the human's (-5.29). This suggests the human-AI system properly learns how to defer tasks to the right decision maker and leverage their respective expertise to achieve human-AI complementarity. 

\begin{table}[!ht]
    \centering
    \resizebox{\linewidth}{!}{
    \begin{tabular}{cccccccc}
    \toprule 
        ~ & DM                          & Human                        & DR-AI                                               & HAI-IPS                       & AIAI       & Predict-then-Collaborate-DR                    & HAI-DR                        \\ \midrule
        Reward & $2.60(0.23)$ & $-5.08(0.33)$ & $2.30(0.22)$  & $2.37(0.22)$ & $2.22(0.26)$ & $2.69^*(0.15)$ & $3.13^*(0.14)$\\ \bottomrule
    \end{tabular}}
    \caption{Results for IHDP Data. $*$ indicates the reward is either significantly better than the Human's reward or the DR-AO's reward at the $\alpha=0.05$ significance level.}
    \label{tab:ihdp}
\end{table}

\begin{table}[!ht]
    \centering
    \begin{tabular}{llll}
    \toprule
        Deferral to Human & 0.03  \\ \midrule
        Deferral to AI & 0.97 \\ \midrule
        Average Human Reward when Deferred to Human & 1.94 \\ \midrule
        Average AI Reward when Deferred to Human & 0.12  \\ \midrule
        Average AI Reward when Deferred to AI & 3.16 \\ \midrule
        Average Human Reward when Deferred to AI & -5.29  \\ \bottomrule
    \end{tabular}
    \caption{Deferral Statistics on IHDP Data.}
    \label{tab:deferral_ihdp}
\end{table}

\subsection{Shopping Assistant}
\label{sec:exp_webshop}

As a third stage of our evaluation, we use an application of an AI shopping assistant with real human responses to validate the proposed method. The WebShop dataset is a simulated e-commerce website environment with 1.18 million real-world products and 12,087 crowd-sourced text instructions, built by \citet{yao2022webshop}. The task of a shopping assistant is to help a customer find and purchase a product that meets certain specifications they are searching for.~\citet{yao2022webshop} constructed the dataset as part of an effort to design AI shopping assistants. In this work, we consider whether a human-AI team may outperform either working alone, and whether the proposed method outperforms the benchmarks. 

The WebShop environment is particularly valuable because it includes predefined rewards and counterfactuals, enabling a rigorous assessment of different methods. By dataset design, we know what the rewards of choosing to purchase each item as a response to a given query would be. Such information is typically difficult or impossible to obtain with observational data, where counterfactuals are inherently missing. By providing a controlled yet realistic task, the dataset allows us to benchmark performance under conditions that closely mimic real-world applications, addressing key challenges in policy evaluation and causal inference.

In the original problem setting, given a text that describes the product requirement, an AI agent needs to navigate different webpages to find, customize, and purchase an item. We simplify the problem and focus on the final purchase stage. As a result, the task is the following: given a text instruction such as ``I'm looking for a small portable folding desk that is already fully assembled; it should have a khaki wood finish, and price lower than 140.00 dollars'', and a webpage with the item description, image, and purchase options, the decision-maker needs to decide whether to purchase the item. \citet{yao2022webshop} collected over 1600 real human demonstrations of the complete shopping assist task. We extract all the events in which a humans need to decide whether to purchase an item or not (which happens once they have navigated to a website), which results in 3025 decisions. In this environment we have access to the optimal item for each text instruction. Thus, we define the potential outcome of choosing to purchase ($T=1$) as $Y_1 = 3 \mathbb{I}(\text{the item is the optimal item}) + (-1)\mathbb{I}(\text{the item is not the optimal item}) + \epsilon$. We define the potential outcome of choosing not to purchase ($T=0$) as $Y_0 = (-1) \mathbb{I}(\text{the item is the optimal item}) + \mathbb{I}(\text{the item is not the optimal item}) + \epsilon$. In both instances, $\epsilon\sim\mathcal{N}(0,0.1)$. Note that $Y_1, Y_0$ represent the reward (e.g., user satisfaction) if the item is purchased or not purchased, respectively. Here, the task of the AI or the human assistant is to decide whether to purchase the item given the text instruction from a user.

The results are shown in \Cref{tab:webshop}. The proposed method yields the best results across all alternatives, and human-AI collaboration indeed yields better performance than either option alone.  
The deferral statistics are shown in \Cref{tab:deferral_webshop}. It is noteworthy that the deferral collaboration system significantly improves with respect to the AI-only baseline with only 7\% of the decision tasks sent to the humans. 
Unlike the previous task, humans are very strong baselines compared to the AI since there are many data modalities and a limited number of samples for this task. 
On the instances that the human is assigned to, humans have a significantly better reward than the AI, and AI has a similar performance compared to humans on tasks assigned to the AI. Thus, the deferral collaboration design could significantly reduce costs of personal shopping assistants, while retaining the quality of human performance by incurring in a small additional cost associated to having humans assess a small percentage (7\%) of the instances.

\begin{table}[!ht]
    \resizebox{\linewidth}{!}{
    \centering
    \begin{tabular}{cccccccc}
    \toprule
        ~ & DM & Human                         & DR-AI                            & HAI-IPS                        & AIAI                                                 & Predict-then-Collaborate-DR & HAI-DR                       \\ \midrule
        webshop & $0.69(0.04)$ & $0.89(0.01)$ & $0.90(0.01)$ & $0.95(0.04)$ &  $0.94(0.04)$ & $0.92(0.02)$ & $0.94^*(0.01)$  \\ \bottomrule
    \end{tabular}}
    \caption{Results for Shopping Assistant Data. $*$ indicates the reward is either significantly better than the Human's reward or the DR-AO's reward at the $\alpha=0.05$ significance level.}
    \label{tab:webshop}
\end{table}

\begin{table}[!ht]
    \centering
    \begin{tabular}{ll}
    \toprule 
        Deferral to Human & 0.07 \\ \midrule
        Deferral to AI & 0.93 \\ \midrule
        Human Reward on Human Assigned Tasks & 1.33 \\ \midrule
        AI Reward on Human Assigned Tasks & 0.41 \\ \midrule
        AI Reward on AI Assigned Tasks & 0.93 \\ \midrule
        Human Reward on AI Assigned Tasks & 0.85 \\ \bottomrule
    \end{tabular}
    \caption{Deferral Statistics on Shopping Assistant Data.}
    \label{tab:deferral_webshop}
\end{table}

\section{Practical Managerial Insight}

Human-AI collaboration offers valuable practical insights for managers navigating AI integration in decision-making processes. The differential strengths as well as the unique dynamics between human and AI decision-making can be strategically leveraged to improve performance outcomes, especially in complex or uncertain environments. This section synthesizes key takeaways and recommendations for managers based on our findings.

\subsection{Diversity Bonus: Develop Complementary AI for Human-AI Teams}

While most work in policy learning focus on developing fully autonomous AI policies, 
one of the central insights of our work is that the AI policy with the highest independent performance may not be the optimal partner in a human-AI collaboration setting in which the work is strategically divided among humans and AI. The type of collaboration approach considered by this work corresponds to a ``divide and conquer" strategy, in which team members collaborate by strategically dividing tasks among them. While the Information Systems literature has recently considered the potential benefits of this type of human-AI collaboration strategy~\citep{fugener2022cognitive}, it has done so by assuming an AI system that is trained independently from the humans to maximize performance over the entire dataset. This diversity bonus emphasizes that an AI system specifically tailored to complement human decision-making can lead to superior team performance compared to either humans or AI working independently. Moreover, the AI teammate should be specifically designed to complement the human counterparts. For instance, while a standard AI might excel across a broader range of tasks, a purpose-built AI teammate, trained to excel, relative to a general-purpose AI, at specific areas where humans tend to underperform, can enhance overall outcomes by strategically filling and excelling at performance gaps.

To capitalize on this and produce the greatest benefits, managers should invest in developing AIs that are not simply high-performing, but that are adapted to the specific strengths and limitations of their human colleagues. Such AIs should be optimized to defer tasks to human decision-makers in scenarios where human intuition or expertise is beneficial and to handle tasks autonomously when their predictive models are likely to yield better outcomes. This approach reflects the principle that “the best individual is not always the best teammate”—an idea supported in cognitive diversity literature and applicable in human-AI collaboration.

\subsection{Effective Collaboration in Imperfect Decision Environments}
Another critical insight is the importance of managing imperfect decision environments, where neither humans nor AIs are flawless across. In these contexts, human-AI collaboration can effectively compensate for mutual weaknesses. Rather than seeking to eliminate all human or AI error, which is often infeasible, the goal is to create a robust collaboration system that intelligently routes decisions to the most suitable agent—human or AI—based on the nature of the task and the historical performance of each agent under similar conditions. Crucially, the methodology proposed in this work unlocks the possibility to build such systems, and to do so using data that organizations routinely collect: historical data of outcomes observed after certain decisions were taken. The empirical results also highlight an important benefit of the proposed system. Human-AI collaboration is attainable in settings where humans exhibit an overall performance that is much inferior to an independent AI, as seen in~\cref{sec:exp_ihdp}, and also in settings where humans are a very strong baseline, as we establish in~\cref{sec:exp_webshop}. %By jointly learning the router and the AI agent, managers can obtain systems that are best positioned to complement humans, striking a division of labor among team members that maximizes expected reward.

\subsection{Strategic Deferral System to Optimize Collaboration}

A key managerial challenge in the integration of AI as part of decision-making pipelines is \emph{how} to design human-AI teams. The human-in-the-loop architecture~\citep{fugener2021will,jussupow2021augmenting}, in which a human who ultimately makes a decision receives an AI recommendation, is arguably the most common human-AI team design. However, as with humans, multiple forms of collaboration among team members are possible and can be advantageous. The potential of collaboration via deferral has recently been studied in both computer science~\citep{madras2018predict,wilder2020learning} and information systems~\citep{fugener2022cognitive}. In this human-AI collaboration design, the workload is smartly distributed among humans and AI to capitalize on the skills of both. However, all work in this space has either assumed the AI is given rather than learned to complement the human and/or has been limited to handling classification tasks, and ill-suited for decision-making scenarios in which outcomes are observed conditioned on the decision made. 

The proposed approach unlocks the possibility to train and deploy deferral systems to handle decision-making tasks. By jointly learning the router and the AI agent, managers can obtain systems that are best positioned to complement humans, striking a division of labor among team members that maximizes expected reward. In such systems, there is a \emph{partial} automation of decision-making, as some instances are handled autonomously by the AI, while other instances are routed to humans. This human-AI collaboration design can offer significant cost-reduction and improved decision performance. The empirical results in~\Cref{sec:exp_ihdp} and ~\cref{sec:exp_webshop} highlight an important advantage in terms of cost-reduction. Across both sets of experiments, involving a human through a deferral collaboration system yields performance that is much better than the AI alone, while only incurring in a small additional cost, since the smart assignment allows the system to significantly improve performance by involving humans in only a small percentage of decisions. 
Such a strategic deferral mechanism could, for example, be integrated into customer service environments, such that some customer complaints are handled independently by AI while others are routed to humans. Similarly, financial decisions, such a loan decisions or investments, could also be handled by human-AI teams via deferral systems proposed in this work.%where AIs manage routine cases independently while deferring complex decisions to human experts.

\section{Conclusions and Future Work}

In this paper, we highlight the potential and challenges inherent in human-AI collaborative decision-making. By proposing a deferral-based framework, we demonstrated that a carefully designed human-AI team composed of an AI collaborator tailored for a given human and a smart task allocation mechanism can be relied on to outperform either humans or AI operating independently. This approach shows that maximizing collaborative outcomes requires designing AI systems that complement human strengths and compensate for their limitations, rather than simply selecting the AI with the highest standalone reward. Our findings underscore the importance of alignment between AI capabilities and human needs within collaborative systems, and provides methodology to achieve this goal. By training AI to defer tasks to human experts in complex cases, we can significantly improve decision quality, especially in areas such as customer service, healthcare, and financial decision-making, where nuanced judgment often outweighs strict algorithmic optimization. Additionally, our theoretical and empirical results suggest that a doubly robust framework is particularly effective in mitigating model misspecifications, providing a more resilient approach to policy learning in human-AI collaboration contexts.

While our work provides an initial attempt in designing the human-AI system, it is not without limitations. First, our framework assumes that both human and AI models remain relatively stable during the collaborative process. However, future research could explore adaptive models that dynamically adjust their routing and decision policies based on real-time feedback, allowing the system to evolve as both human and AI capabilities change over time. Future work can consider an online learning system \citep{yan2018active,jesson2021causal,cao2024hr} that adaptively learn human decisions. Second, we restrict ourselves in the deferral collaboration framework in this paper and there may be other forms of collaborative systems. Future work can explore other forms of human-AI collaborative systems in policy learning. Last, it is essential that AI deferral systems are designed to mitigate biases that might disproportionately affect specific groups. Future work could explore fairness-aware deferral mechanisms that account for potential inequities in routing decisions. This may involve incorporating fairness constraints into routing algorithms or developing monitoring tools to detect biases over time. 

\newpage
\bibliographystyle{informs2014}
\bibliography{informs2014}

%\iffalse
\newpage 

\begin{APPENDICES}
\crefalias{section}{appendix}

\section{Proofs}\label{app:proof}

Proof of \Cref{prop:dr}: 

\proof{Proof:}
When $f$ is consistent, 
\begin{align}
\mathbb{E}\hat{\theta}_{\text{DR}} = \mathbb{E} \phi(X)(Y - C(X))  + (1-\phi(X))\sum_T \pi(T|X)Y(X,T) = \theta(\phi,\pi).
\end{align}
When $\hat{\pi}_0$ is consistent, 
\begin{align}
\mathbb{E}\hat{\theta}_{\text{DR}} = \mathbb{E} \phi(X)(Y - C(X))  + \mathbb{E}_{\pi_0}(1-\phi(X))  \frac{\pi(T|X)}{\pi_0(T|X)}Y(X,T) = \theta(\phi,\pi).
\end{align}
    $\blacksquare$
\endproof 

Proof of \Cref{lemma:tildedelta}:

\proof{Proof:}

By the symmetrization argument 
\begin{align}
       &  \sup_{\substack{\pi_1,\pi_2 \in\Pi\\ \phi_1,\phi_2\in\Phi}}
|\Delta(\{\pi_1,\phi_1\},\{\pi_2,\phi_2\})-\tilde{\Delta}(\{\pi_1,\phi_1\},\{\pi_2,\phi_2\})| \nonumber  \\
& \leq 2 \mathbb{E}\sup\frac{1}{n}|\sum_{i=1}^n Z_i 
\Big(
\langle \pi_1-\pi_2,\Gamma_i\rangle + 
\langle \phi_1-\phi_2,Y_i\rangle +
\langle \pi_2,\Gamma_i\rangle\phi_2(X_i) - 
\langle \pi_1,\Gamma_i\rangle\phi_1(X_i)
\Big)| \\ 
& \leq 2 \mathbb{E}\sup\frac{1}{n}|\sum_{i=1}^n Z_i \langle \pi_1-\pi_2,\Gamma_i\rangle| + 
2 \mathbb{E}\sup\frac{1}{n}|\sum_{i=1}^n Z_i \langle \phi_1-\phi_2,Y_i\rangle| + 
2 \mathbb{E}\sup \frac{1}{n}|\sum_{i=1}^n Z_i\langle \pi,\Gamma_i\rangle| \nonumber  \\ 
& \leq 54.4\sqrt{2}(\textit{k}(\Pi)+8)\sqrt{\frac{\sup_{\pi_1,\pi_2\in\Pi }\mathbb{E}[\langle\Gamma_i,\pi_1(X_i)-\pi_2(X_i)\rangle^2]}{n}} 
 + 27.2\sqrt{2}(\textit{k}(\Pi)+8)\sqrt{\frac{\sup_{\pi\in\Pi }\mathbb{E}[\langle\Gamma_i,\pi(X_i)\rangle^2]}{n}}
\nonumber \\
& \quad + 54.4\sqrt{2}(\textit{k}(\Phi)+8)\sqrt{\frac{\sup_{\phi_1,\phi_2\in\Phi }\mathbb{E}[\langle Y_i,\phi_1(X_i)-\phi_2(X_i)\rangle^2]}{n}} + o(\frac{1}{\sqrt{n}}) \nonumber \\
& \leq 71.6 \sqrt{2}(\textit{k}(\Pi)+8)\sqrt{\frac{\sup_{\pi_1,\pi_2\in\Pi }\mathbb{E}[\langle\Gamma_i,\pi_1(X_i)-\pi_2(X_i)\rangle^2]}{n}} \nonumber \\
& \quad + 54.4\sqrt{2}(\textit{k}(\Phi)+8)\sqrt{\frac{\sup_{\phi_1,\phi_2\in\Phi }\mathbb{E}[\langle Y_i,\phi_1(X_i)-\phi_2(X_i)\rangle^2]}{n}} + o(\frac{1}{\sqrt{n}})
\end{align}

The second inequality is from the contraction principle. The third inequality is a rearrangement of Lemma 1. 
This indicates the regret is worse than \citet{zhou2023offline} only in the constant since our policies are more complex because of the router and the human policy. 
$\blacksquare$
\endproof

Proof of \Cref{lemma_hat_delta_rate}:

\proof{Proof:}
With some algebra, we have 
\begin{align}
    & \hat{\Delta}^j(\{\pi_1,\phi_1\},\{\pi_2,\phi_2\})-\tilde{\Delta}^j(\{\pi_1,\phi_1\},\{\pi_2,\phi_2\}) \nonumber \\ 
    & = \frac{1}{n}\sum_{i=1}^n (\pi_1^j(X_i)-\pi_2^j(X_i)+\pi_2^j(X_i)\phi_2(X_i)-\pi_1^j(X_i)\phi_1(X_i))(\hat{f}^{-k(i)}(X_i,a^j)-{f}^{-k(i)}(X_i,a^j))(1-\frac{\mathbb{I}[A_i=a^j]}{{\pi}_0(a^j|X_i)}) \nonumber \\
    & + \frac{1}{n}\sum_{A_i=a^j} (\pi_1^j(X_i)-\pi_2^j(X_i)+\pi_2^j(X_i)\phi_2(X_i)-\pi_1^j(X_i)\phi_1(X_i))
    (Y_i-f(X_i,a^j))(\frac{1}{\hat{\pi}_0^{-k(i)}(a^j|X_i)} - \frac{1}{\pi_0(a^j|X_i)}) \nonumber \\
    & + \frac{1}{n}\sum_{A_i=a^j} (\pi_1^j(X_i)-\pi_2^j(X_i)+\pi_2^j(X_i)\phi_2(X_i)-\pi_1^j(X_i)\phi_1(X_i)) \nonumber \\
    & \qquad \qquad\qquad\times (\frac{1}{\hat{\pi}_0^{-k(i)}(a^j|X_i)} - \frac{1}{\pi_0(a^j|X_i)})
    (\hat{f}^{-k(i)}(X_i,a^j)-{f}^{-k(i)}(X_i,a^j)) \nonumber 
\end{align}

Since the first and the second term are the product and estimation error and noise and the last term is the product of two noise term, 
we can show the first, second and the third term are all $o_p(\frac{1}{\sqrt{n}})$ following the standard arguments in \citet{zhou2023offline}, which concludes the proof. 
$\blacksquare$
\endproof

Proof of \Cref{thm:main}:

\proof{Proof:}

First, we can write 
\begin{align}
R(\hat{\pi},\hat{\phi}) & = \hat{\theta}(\pi^*,\phi^*)-\hat{\theta}(\hat{\pi},\hat{\phi}) + \Delta(\{\pi^*,\phi^*\},\{\hat{\pi},\hat{\phi}\})    - \hat{\Delta}(\{\pi^*,\phi^*\},\{\hat{\pi},\hat{\phi}\}) \\
&\leq \Delta(\{\pi^*,\phi^*\},\{\hat{\pi},\hat{\phi}\})    - \hat{\Delta}(\{\pi^*,\phi^*\},\{\hat{\pi},\hat{\phi}\}) \\ 
&\leq \sup_{\substack{\pi_1,\pi_2 \in\Pi\\ \phi_1,\phi_2\in\Phi}}|
\Delta(\{\pi_1,\phi_1\},\{\pi_2,\phi_2\})    - \hat{\Delta}(\{\pi_1,\phi_1\},\{\pi_2,\phi_2\})| \\ 
& \leq 
\sup_{\substack{\pi_1,\pi_2 \in\Pi\\ \phi_1,\phi_2\in\Phi}}
|\Delta(\{\pi_1,\phi_1\},\{\pi_2,\phi_2\})-\tilde{\Delta}(\{\pi_1,\phi_1\},\{\pi_2,\phi_2\})| \nonumber  \\
& \quad  + 
\sup_{\substack{\pi_1,\pi_2 \in\Pi\\ \phi_1,\phi_2\in\Phi}}
|\hat{\Delta}(\{\pi_1,\phi_1\},\{\pi_2,\phi_2\})-\tilde{\Delta}(\{\pi_1,\phi_1\},\{\pi_2,\phi_2\})|
\end{align}

Next we need to bound both terms separately. Define $V^* = \max\{\sup_{\phi_1,\phi_2\in\Phi }\mathbb{E}[\langle Y_i,\phi_1(X_i)-\phi_2(X_i)\rangle^2], \sup_{\phi_1,\phi_2\in\Phi }\mathbb{E}[\langle Y_i,\phi_1(X_i)-\phi_2(X_i)\rangle^2] \}$, and applying Lemma 2 and 3 finishes the proof. 
$\blacksquare$
\endproof

\section{No Overlap}\label{app:overlap}
Here we discuss the setting where the overlap is violated. The overlap condition implies that all possible actions in the historical data will be selected with a positive probability, which is very likely to hold for human-generated data due to the inherent uncertainty in human decisions \citep{karlinsky2019automating}. When the overlap assumption is violated, the theoretical analysis would fail and here we introduce the mitigation strategy. 

Denote the set of instances where the overlap condition does not hold and hold as $\mathcal{NO}$ and $\mathcal{O}$, respectively. We can train our proposed algorithms in the set $\mathcal{O}$, and adopt the deterministic human decisions in the set $\mathcal{NO}$. For the cases in $\mathcal{NO}$ where we have clear rationale of why humans take determistic actions, the rationale can be complied as a rule (or learned by an additional machine learning model) and be fully automated in the future. The performance of a future human-AI system under $\pi,\phi$ can be estimated by 

{\small
 \begin{align}
& \hat{\theta}_{\text{NO-DR}} = 
\sum_{X_i\in\mathcal{NO}} Y_i + 
\nonumber\\ 
& \sum_{X_i\in\mathcal{O}} \phi(X_i)(Y_i - C(X_i))  + (1-\phi(X_i)) \Big(\frac{\pi(T_i|X_i)}{\hat{\pi}_{0}^{-k(i)}(T_i|X_i)}(Y_i-f^{-k(i)}(X_i,T_i)) + \sum_T \pi(T|X_i)f^{-k(i)}(X_,T) \Big). 
\end{align}
}

Note that the first term is not dependent on $\pi,\phi$, so optimize this system is equivalent to replacing the full dataset with $\mathcal{O}$ in the main paper. 

\section{Additional Experiments}\label{app:addexp}

In this section, following the common practie in policy learning literatures \citep{zhan2024policy,swaminathan2015counterfactual}, we adapt multiclassification datasets to policy learning problems by
treating each class as an arm and the one-hot encoding of the associated label as the counterfactual outcome of different arms. More specifically, for a classification problem, the reward will be 1 if the treatment equals the corresponding class label. For example, if the data instance has a label of 0, its potential outcome for treatment 0 and 1 will be 1 and 0, respectively.  

We use two datasets Adult \citep{adult_2} and Credit \citep{statlog_(german_credit_data)_144} in the experiments. To simulate human behaviors, we assume human will select the optimal actions with probability $0.9$ when $\beta^\top x > Q(\beta^\top x, 0.5)$, and with probability $0.2$ when $\beta^\top x \leq Q(\beta^\top x, 0.5)$, where each entry of $\beta$ is sampled from the uniform distribution $U(0,1)$. Since we have demonstrated that the non-doubly-robust methods are fragile to model misspecification in the main paper, here we only compare the doubly-robust variants. We use gradient boosted tree as the direct method with the default parameters in the experiments. 

The results for the average reward and the deferral statistics are shown in \Cref{app:clfdata_reward} and \Cref{app:clfdata_stat}, respectively. The proposed HAI-DR method often has significant better reward compared to the human-ony and the DR-AI baseline. Interestingly, we find that the deferral rate is around 50\% and the average reward is around 0.9, which are exactly the proportion of the instances the humans are good at and the human average reward on these instances, respectively. This suggests the proposed method correctly identifies the subset of instances that humans are good at and defer these tasks to humans. 

\begin{table}[!ht]
    \centering
    \begin{tabular}{ccccc}
    \toprule
         & Human & DR-AI & Predict-then-Collaborate-DR & HAI-DR \\ \midrule
        Adult & $0.55(0.01)$ & $0.52(0.03)$ & $0.65^*(0.02)$ & $0.66^*(0.03)$ \\ 
        Credit & $0.56(0.00)$ & $0.39(0.03)$ & $0.55(0.02)$ & $0.60(0.02)$ \\ 
        Kin8nm & $0.55(0.00)$ & $0.30(0.01)$ & $0.58(0.04)$ & $0.61^*(0.01)$ \\ \bottomrule
    \end{tabular}
    \caption{Reward on Different Datasets. $*$ indicates the reward is significantly better than both the Human's reward and the DR-AI's reward at the $\alpha=0.05$ significance level.}
    \label{app:clfdata_reward}
\end{table}

\begin{table}[!ht]
    \centering
    \begin{tabular}{cccc}
    \toprule
        ~ & Adult & Credit & Kin8nm \\ \midrule
        Deferral to Human & 0.50 & 0.45 & 0.50 \\ 
        Deferral to AI  & 0.50 & 0.55 & 0.50 \\ 
        Human Reward on Human Assigned Tasks & 0.88 & 0.81 & 0.90 \\ 
        AI Reward on Human Assigned Tasks  & 0.48 & 0.40 & 0.31 \\ 
        AI Reward on AI Assigned Tasks  & 0.44 & 0.43 & 0.33 \\ 
        Human Reward on AI Assigned Tasks & 0.22 & 0.35 & 0.21 \\ \bottomrule
    \end{tabular}
    \caption{Deferral Statistics on Different Datasets}
    \label{app:clfdata_stat}
\end{table}

\end{APPENDICES}
%\fi 

%%%%%%%%%%%%%%%%%
\end{document}

%% file: intro_r1_revised.tex
The field of management has long grappled with the problem of strategic task allocation, which seeks to coordinate the efforts of multiple actors to achieve organizational goals. This classical paradigm, framed as a problem of optimization, has been applied to diverse contexts, from allocating work between machines in distributed manufacturing systems \citep{choy2000task,ernst2006exact} to distributing tasks among humans~\citep{sridhar1997incomplete,lindberg2024entrainment,hassin2021self,manshadi2020online}. 
Generally speaking, strategic cost allocation aims to maximize overall utility by strategically allocating tasks to different actors, each of which may have different skills and associated costs. In the context of decision-making tasks, strategic task allocation aims to assign each instance to the decision-maker most likely to make a correct decision for a given instance. 

The introduction of artificial intelligence (AI) as a decision-maker presents a paradigm shift in strategic task allocation for decision-making. The skills of the different agents have typically been thought of as fixed, thus making the optimization problem solely about allocation. However, the AI is not merely another fixed agent to which tasks can be allocated; it is a partner that can be designed to form an optimal, synergistic team with its human counterparts. 
An AI agent trained considering human agent behavior can yield better human-AI team performance than the best independent AI, who may share the same strengths or weaknesses of the human decision maker thus limiting augmentation \citep{bansal2020optimizing}. 
This fundamentally enriches the optimization problem, as the task is now to \emph{jointly} learn both the AI decision-maker and the strategic allocation. 
This task, sometimes referred to as deferral collaboration~\citep{madras2018predict} or delegation~\citep{fugener2022cognitive}, aims to create a synergistic human-AI partnership that exceeds what either humans or AI could accomplish independently, ultimately aiming to maximize human-AI complementarity~\citep{wilder2020learning}.

Crucially, existing research on deferral collaboration has only considered the task of supervised classification~\citep{madras2018predict,wilder2020learning,fugener2022cognitive}. In this setting, the objective is to predict a correct label $Y$, for a given instance $X$, using training data where ground-truth labels are available. For example, in developing a tool for medical diagnostics, a model may be trained on a dataset where a ``gold standard'' diagnosis is known for all patients \citep{wilder2020learning}. When optimizing for complementarity, this availability of ground truth allows for a direct comparison between human predictions $Y_H$, AI predictions $\hat{Y}$, and the gold standard label $Y$, making it straightforward to identify each partner's relative strengths and weaknesses. However, this reliance on ground-truth labels renders existing methods inapplicable to a vast range of critical managerial decision-making tasks.
Concrete, practical examples include customer service, where agents must select the best response to improve satisfaction; personalized advertising, where a platform must select content to maximize engagement; 
and precision medicine, where clinicians must choose the optimal treatment for a patient. In all these settings, the goal is to select the most advantageous action (policy). In available observational data, however, outcomes are only observed for decisions or treatments $T$ that were chosen in the past. Such settings yield data with missing counterfactuals, resulting in a task that corresponds to policy learning rather than standard supervised learning \citep{athey2017efficient,zhou2023offline}. 
%These tasks stand to benefit greatly from human-AI deferral collaboration. In scenarios where one agent performs suboptimally, strategically deferring the decision to another complementary agent may improve overall decision quality and mitigate the risks of errors. 
While policy learning for fully autonomous decision-making has received substantial attention \citep{swaminathan2015counterfactual,kallus2018balanced,si2020distributionally}, the problem of human-AI complementarity in this domain remains largely unexplored. This is a significant gap in the literature, as these tasks stand to benefit greatly from human-AI deferral collaboration by enabling decision performance that is better than either humans or AI alone.

%such as customer service, personalized pricing, and precision medicine, in which the goal is to select the most advantageous action that yields the best outcome, but  where outcomes are only observed conditioned on decisions or treatments that had been actually taken, $T$. Such settings yield observational data with missing counterfactuals, in which case the task is not supervised learning but policy learning \citep{athey2017efficient,zhou2023offline}. While policy learning for autonomous decision-making has received substantial attention \citep{swaminathan2015counterfactual,kallus2018balanced,si2020distributionally}, the problem of human-AI complementarity in this domain remains largely unexplored.

\begin{figure}
    \centering
    \includegraphics[width=0.8\linewidth]{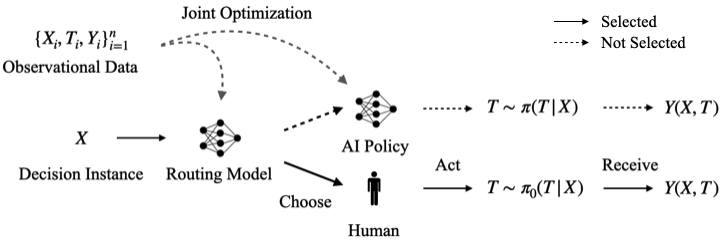}
    \caption{Learning Complementary Policies for Human-AI augmentation \textsc{lcp-hai}. The routing model and AI policy are jointly optimized on the observational data. After the Human-AI system is deployed, the routing model will select between AI and the human to solve the task. The figure demonstrates a case where a human decision maker is chosen to solve the task, the human then selects a treatment and the requester receives a reward under the assigned treatment.}
    \label{fig:dag}
\end{figure}

\textbf{To the best of our knowledge, ours is the first work to propose a deferral collaboration model for policy learning}. Figure~\ref{fig:dag} outlines the deferral collaboration framework we consider. As visualized in the diagram, each decision instance can be routed to either an AI algorithm (which ought to be learned as well) or a human decision maker, and we only observe the outcome under the assigned treatment. In this work we propose a method that \textit{jointly} learns an AI algorithm that aims to complement the human, which we refer to as the \emph{AI policy}, and a routing model, which advantageously routes tasks. These models are learned using observational data of human decisions and corresponding outcomes. 
The overall goal of the joint learning is to maximize the benefits of human-AI complementarity. This is achieved by \emph{simultaneously} learning two components: (1) an AI policy that is designed to be most complementary to the human, and (2) a routing model that assigns each task to the most suitable partner.
We refer to this problem as Learning Complementary Policies for Human-AI Augmentation (\textsc{lcp-hai}).

As an illustrative example, consider the task of responding to customer complaints. In historical data available for learning, one would observe human agents' decisions of how to respond to customer complaints. For instance, if an agent offers flight miles (the treatment $T$), one would observe the resulting customer satisfaction, $Y(X,T=\text{flight miles})$. However, the counterfactual outcome---what the satisfaction would have been had the agent offered a simple apology $Y(X,T=\text{apology})$---is not observed. When learning a complementary policy for human-AI augmentation, the goal is to learn an AI that can autonomously handle some of the complaints comparably or better than the humans, while deferring to humans complaints that they would handle better. The AI decision-maker and the routing model to determine who to route each complaint to are jointly learned. The AI learned is thus dependent on the skills of the humans it will complement.

%Since humans and AI may have different decision making behaviors, the task-focuses, reward-maximizing AI-policy---i.e., 
Training an AI that is specifically designed to \emph{complement} humans is essential to maximizing human-AI augmentation, and can yield substantially better results than using an AI policy trained to maximize the total reward when acting alone. 
%on the underlying task when the AI makes all the decisions by itself---may not be the same as the complementary AI policy that maximizes reward in a deferral collaboration context in which it collaborates with a given human or group of humans. 
We will explore this question carefully in Section~\ref{sec:best_vs_complementary}, but the intuition is simple and familiar to business scholars. As \citet{page2019diversity} notes, the best human teams are the ones in which members complement one another; for example, the best tennis player is not necessarily the best doubles teammate~\citep{bansal2021does}. Given the AI policy is not given but learned, producing the AI policy that exhibits the best complementarities can be advantageous: 
although the resulting AI collaborator might exhibit lower overall performance compared to a purely reward-maximizing counterpart, it may yield significantly better outcomes when tasks are strategically divided between AI and humans, because there is an opportunity to specialize on instances that are challenging to humans. The possibility for complementarity is further amplified when considering that the AI algorithm is jointly optimized with the routing algorithm that allocates instances to the AI or to a human, which expands the space of possible human-AI team arrangements considered. % to allow the Ai collaborator complements the human weaknesses. 
%This furthers the possibility for complementarity, because the routing model, which determines how to strategically divide the work, is optimized jointly with the AI policy.

The fundamental problem in finding a good deferral collaboration policy is how to evaluate a potential human-AI policy candidate using the observational data, in which we only observe outcomes for the assigned treatments. 
In this paper, we develop a novel deferral collaboration Doubly-Robust (DR) estimator \citep{athey2017efficient,zhou2023offline} to optimize a human-AI deferral collaboration system. We provide theoretical and empirical evidence showing that the proposed approach is better than existing alternatives that could be used to account for the lack of counterfactual outcomes. 

Below we summarize the core contributions of our work:

\begin{itemize}
    \item We formulate the problem of Learning Complementary Policies for Human-AI augmentation \textsc{lcp-hai}. The goal is to leverage human-machine complementarity so as to achieve maximum total reward through a deferral collaboration system in settings where only outcomes under assigned decisions are observed. 
    \item We propose a \textsc{lcp-hai} method. To the best of our knowledge, ours is the first method that tackles this problem.  
    Using observational data of human decisions, we jointly train a router and an AI policy to best exploit the AI and the humans' decision-making abilities. The resulting system can route new instances to either a human decision-maker or to the AI, which has been trained to best complement humans.
    \item We theoretically demonstrate that the proposed method is doubly-robust when either the reward model or the human behavior model is misspecified. We also show that the proposed method has faster convergence properties compared to learning either model alone. 
    \item We empirically study the performance of the proposed solutions using synthetic, semi-synthetic, and real human decisions. 
    Our results demonstrate that our approach yields significant improvement over baselines and that these benefits are achieved by both effectively  \emph{learning} and \emph{leveraging}  human-AI decision-making complementarity.
    \item We discuss managerial implications of our research for a wide array of impactful business problems and  discuss how our approach and the principles it is based on offer promising foundations on which future work on human-AI augmentation can build upon.
\end{itemize}

%% file: arxiv_v1.bbl
\begin{thebibliography}{55}
\providecommand{\natexlab}[1]{#1}
\providecommand{\url}[1]{\texttt{#1}}
\providecommand{\urlprefix}{URL }

\bibitem[{Athey et~al.(2017)Athey, Wager et~al.}]{athey2017efficient}
Athey S, Wager S, et~al. (2017) Efficient policy learning. Technical report.

\bibitem[{Bansal et~al.(2020)Bansal, Nushi, Kamar, Horvitz, \protect\BIBand{} Weld}]{bansal2020optimizing}
Bansal G, Nushi B, Kamar E, Horvitz E, Weld DS (2020) Optimizing ai for teamwork. \emph{arXiv:2004.13102} .

\bibitem[{Bansal et~al.(2021)Bansal, Wu, Zhou, Fok, Nushi, Kamar, Ribeiro, \protect\BIBand{} Weld}]{bansal2021does}
Bansal G, Wu T, Zhou J, Fok R, Nushi B, Kamar E, Ribeiro MT, Weld D (2021) Does the whole exceed its parts? the effect of ai explanations on complementary team performance. \emph{Proceedings of the 2021 CHI Conference on Human Factors in Computing Systems}, 1--16.

\bibitem[{Becker \protect\BIBand{} Kohavi(1996)}]{adult_2}
Becker B, Kohavi R (1996) {Adult}. UCI Machine Learning Repository, {DOI}: https://doi.org/10.24432/C5XW20.

\bibitem[{Bertsimas \protect\BIBand{} Kallus(2020)}]{bertsimas2020predictive}
Bertsimas D, Kallus N (2020) From predictive to prescriptive analytics. \emph{Management Science} 66(3):1025--1044.

\bibitem[{Bondi et~al.(2022)Bondi, Koster, Sheahan, Chadwick, Bachrach, Cemgil, Paquet, \protect\BIBand{} Dvijotham}]{bondi2022role}
Bondi E, Koster R, Sheahan H, Chadwick M, Bachrach Y, Cemgil T, Paquet U, Dvijotham K (2022) Role of human-ai interaction in selective prediction. \emph{Proceedings of the AAAI Conference on Artificial Intelligence}, volume~36, 5286--5294.

\bibitem[{Cao et~al.(2024)Cao, Gao, \protect\BIBand{} Keyvanshokooh}]{cao2024hr}
Cao J, Gao R, Keyvanshokooh E (2024) Hr-bandit: Human-ai collaborated linear recourse bandit. \emph{arXiv preprint arXiv:2410.14640} .

\bibitem[{Chernozhukov et~al.(2017)Chernozhukov, Chetverikov, Demirer, Duflo, Hansen, Newey, \protect\BIBand{} Robins}]{chernozhukov2017double}
Chernozhukov V, Chetverikov D, Demirer M, Duflo E, Hansen C, Newey W, Robins J (2017) Double/debiased machine learning for treatment and causal parameters. Technical report.

\bibitem[{Choy \protect\BIBand{} Lee(2000)}]{choy2000task}
Choy K, Lee W (2000) Task allocation using case-based reasoning for distributed manufacturing systems. \emph{Logistics Information Management} 13(3):167--176.

\bibitem[{De et~al.(2020)De, Koley, Ganguly, \protect\BIBand{} Gomez-Rodriguez}]{de2020regression}
De A, Koley P, Ganguly N, Gomez-Rodriguez M (2020) Regression under human assistance. \emph{AAAI}, 2611--2620.

\bibitem[{Dud{\'\i}k et~al.(2014)Dud{\'\i}k, Erhan, Langford, Li et~al.}]{dudik2014doubly}
Dud{\'\i}k M, Erhan D, Langford J, Li L, et~al. (2014) Doubly robust policy evaluation and optimization. \emph{Statistical Science} 29(4):485--511.

\bibitem[{Dudley(1967)}]{dudley1967sizes}
Dudley RM (1967) The sizes of compact subsets of hilbert space and continuity of gaussian processes. \emph{Journal of Functional Analysis} 1(3):290--330.

\bibitem[{D’Amour et~al.(2021)D’Amour, Ding, Feller, Lei, \protect\BIBand{} Sekhon}]{d2021overlap}
D’Amour A, Ding P, Feller A, Lei L, Sekhon J (2021) Overlap in observational studies with high-dimensional covariates. \emph{Journal of Econometrics} 221(2):644--654.

\bibitem[{Elmachtoub \protect\BIBand{} Grigas(2017)}]{elmachtoub2017smart}
Elmachtoub AN, Grigas P (2017) Smart" predict, then optimize". \emph{arXiv:1710.08005} .

\bibitem[{Elmachtoub et~al.(2021)Elmachtoub, Gupta, \protect\BIBand{} Hamilton}]{elmachtoub2021value}
Elmachtoub AN, Gupta V, Hamilton ML (2021) The value of personalized pricing. \emph{Management Science} 67(10):6055--6070.

\bibitem[{Ernst et~al.(2006)Ernst, Jiang, \protect\BIBand{} Krishnamoorthy}]{ernst2006exact}
Ernst A, Jiang H, Krishnamoorthy M (2006) Exact solutions to task allocation problems. \emph{Management science} 52(10):1634--1646.

\bibitem[{F{\"u}gener et~al.(2021)F{\"u}gener, Grahl, Gupta, \protect\BIBand{} Ketter}]{fugener2021will}
F{\"u}gener A, Grahl J, Gupta A, Ketter W (2021) Will humans-in-the-loop become borgs? merits and pitfalls of working with ai. \emph{Management Information Systems Quarterly (MISQ)-Vol} 45.

\bibitem[{F{\"u}gener et~al.(2022)F{\"u}gener, Grahl, Gupta, \protect\BIBand{} Ketter}]{fugener2022cognitive}
F{\"u}gener A, Grahl J, Gupta A, Ketter W (2022) Cognitive challenges in human--artificial intelligence collaboration: Investigating the path toward productive delegation. \emph{Information Systems Research} 33(2):678--696.

\bibitem[{Gao et~al.(2021)Gao, Biggs, Sun, \protect\BIBand{} Han}]{gao2021enhancing}
Gao R, Biggs M, Sun W, Han L (2021) Enhancing counterfactual classification via self-training. \emph{arXiv preprint arXiv:2112.04461} .

\bibitem[{Hassin \protect\BIBand{} Nathaniel(2021)}]{hassin2021self}
Hassin R, Nathaniel A (2021) Self-selected task allocation. \emph{Manufacturing \& Service Operations Management} 23(6):1669--1682.

\bibitem[{Hill(2011)}]{hill2011bayesian}
Hill JL (2011) Bayesian nonparametric modeling for causal inference. \emph{Journal of Computational and Graphical Statistics} 20(1):217--240.

\bibitem[{Hofmann(1994)}]{statlog_(german_credit_data)_144}
Hofmann H (1994) {Statlog (German Credit Data)}. UCI Machine Learning Repository, {DOI}: https://doi.org/10.24432/C5NC77.

\bibitem[{Jesson et~al.(2021)Jesson, Tigas, van Amersfoort, Kirsch, Shalit, \protect\BIBand{} Gal}]{jesson2021causal}
Jesson A, Tigas P, van Amersfoort J, Kirsch A, Shalit U, Gal Y (2021) Causal-bald: Deep bayesian active learning of outcomes to infer treatment-effects from observational data. \emph{Advances in Neural Information Processing Systems} 34:30465--30478.

\bibitem[{Johansson et~al.(2016)Johansson, Shalit, \protect\BIBand{} Sontag}]{johansson2016learning}
Johansson F, Shalit U, Sontag D (2016) Learning representations for counterfactual inference. \emph{ICML}, 3020--3029.

\bibitem[{Jussupow et~al.(2021)Jussupow, Spohrer, Heinzl, \protect\BIBand{} Gawlitza}]{jussupow2021augmenting}
Jussupow E, Spohrer K, Heinzl A, Gawlitza J (2021) Augmenting medical diagnosis decisions? an investigation into physicians’ decision-making process with artificial intelligence. \emph{Information Systems Research} 32(3):713--735.

\bibitem[{Kallus(2018)}]{kallus2018balanced}
Kallus N (2018) Balanced policy evaluation and learning. \emph{Advances in neural information processing systems} 31.

\bibitem[{Kallus \protect\BIBand{} Zhou(2018)}]{kallus2018confounding}
Kallus N, Zhou A (2018) Confounding-robust policy improvement. \emph{NeurIPS}, 9269--9279.

\bibitem[{Kallus \protect\BIBand{} Zhou(2021)}]{kallus2021minimax}
Kallus N, Zhou A (2021) Minimax-optimal policy learning under unobserved confounding. \emph{Management Science} 67(5):2870--2890.

\bibitem[{Karlinsky-Shichor \protect\BIBand{} Netzer(2019)}]{karlinsky2019automating}
Karlinsky-Shichor Y, Netzer O (2019) Automating the b2b salesperson pricing decisions: Can machines replace humans and when. Available at SSRN:3368402.

\bibitem[{Kitagawa \protect\BIBand{} Tetenov(2018)}]{kitagawa2018should}
Kitagawa T, Tetenov A (2018) Who should be treated? empirical welfare maximization methods for treatment choice. \emph{Econometrica} 86(2):591--616.

\bibitem[{K{\"u}nzel et~al.(2019)K{\"u}nzel, Sekhon, Bickel, \protect\BIBand{} Yu}]{kunzel2019metalearners}
K{\"u}nzel SR, Sekhon JS, Bickel PJ, Yu B (2019) Metalearners for estimating heterogeneous treatment effects using machine learning. \emph{Proceedings of the national academy of sciences} 116(10).

\bibitem[{Langford et~al.(2008)Langford, Strehl, \protect\BIBand{} Wortman}]{langford2008exploration}
Langford J, Strehl A, Wortman J (2008) Exploration scavenging. \emph{Proceedings of the 25th international conference on Machine learning}, 528--535.

\bibitem[{Lindberg et~al.(2024)Lindberg, Schecter, Berente, Hennel, \protect\BIBand{} Lyytinen}]{lindberg2024entrainment}
Lindberg A, Schecter A, Berente N, Hennel P, Lyytinen K (2024) The entrainment of task allocation and release cycles in open source software development. \emph{MIS Quarterly} 48(1):67--94.

\bibitem[{Louizos et~al.(2017)Louizos, Shalit, Mooij, Sontag, Zemel, \protect\BIBand{} Welling}]{louizos2017causal}
Louizos C, Shalit U, Mooij JM, Sontag D, Zemel R, Welling M (2017) Causal effect inference with deep latent-variable models. \emph{Advances in neural information processing systems} 30.

\bibitem[{Madras et~al.(2018)Madras, Pitassi, \protect\BIBand{} Zemel}]{madras2018predict}
Madras D, Pitassi T, Zemel R (2018) Predict responsibly: improving fairness and accuracy by learning to defer. \emph{NeurIPS} 31:6147--6157.

\bibitem[{Manshadi \protect\BIBand{} Rodilitz(2020)}]{manshadi2020online}
Manshadi V, Rodilitz S (2020) Online policies for efficient volunteer crowdsourcing. \emph{Proceedings of the 21st ACM Conference on Economics and Computation}, 315--316.

\bibitem[{Page(2019)}]{page2019diversity}
Page SE (2019) \emph{The diversity bonus: How great teams pay off in the knowledge economy} (Princeton University Press).

\bibitem[{Parikh et~al.(2022)Parikh, Varjao, Xu, \protect\BIBand{} Tchetgen}]{parikh2022validating}
Parikh H, Varjao C, Xu L, Tchetgen ET (2022) Validating causal inference methods. \emph{International conference on machine learning}, 17346--17358 (PMLR).

\bibitem[{Qi et~al.(2023)Qi, Miao, \protect\BIBand{} Zhang}]{qi2023proximal}
Qi Z, Miao R, Zhang X (2023) Proximal learning for individualized treatment regimes under unmeasured confounding. \emph{Journal of the American Statistical Association} 1--14.

\bibitem[{Raghu et~al.(2019)Raghu, Blumer, Corrado, Kleinberg, Obermeyer, \protect\BIBand{} Mullainathan}]{raghu2019algorithmic}
Raghu M, Blumer K, Corrado G, Kleinberg J, Obermeyer Z, Mullainathan S (2019) The algorithmic automation problem: Prediction, triage, and human effort. \emph{arXiv:1903.12220} .

\bibitem[{Rubin(1980)}]{rubin1980randomization}
Rubin DB (1980) Randomization analysis of experimental data: The fisher randomization test comment. \emph{Journal of the American statistical association} 75(371):591--593.

\bibitem[{Rubin(2005)}]{rubin2005causal}
Rubin DB (2005) Causal inference using potential outcomes: Design, modeling, decisions. \emph{JASA} 100(469):322--331.

\bibitem[{Shalit et~al.(2017)Shalit, Johansson, \protect\BIBand{} Sontag}]{shalit2017estimating}
Shalit U, Johansson FD, Sontag D (2017) Estimating individual treatment effect: generalization bounds and algorithms. \emph{ICML}, 3076--3085 (PMLR).

\bibitem[{Si et~al.(2020)Si, Zhang, Zhou, \protect\BIBand{} Blanchet}]{si2020distributionally}
Si N, Zhang F, Zhou Z, Blanchet J (2020) Distributionally robust policy evaluation and learning in offline contextual bandits. \emph{International Conference on Machine Learning}, 8884--8894 (PMLR).

\bibitem[{Sridhar \protect\BIBand{} Balachandran(1997)}]{sridhar1997incomplete}
Sridhar SS, Balachandran BV (1997) Incomplete information, task assignment, and managerial control systems. \emph{Management Science} 43(6):764--778.

\bibitem[{Sulek et~al.(1995)Sulek, Lind, \protect\BIBand{} Marucheck}]{sulek1995impact}
Sulek JM, Lind MR, Marucheck AS (1995) The impact of a customer service intervention and facility design on firm performance. \emph{Management Science} 41(11):1763--1773.

\bibitem[{Swaminathan \protect\BIBand{} Joachims(2015{\natexlab{a}})}]{swaminathan2015counterfactual}
Swaminathan A, Joachims T (2015{\natexlab{a}}) Counterfactual risk minimization: Learning from logged bandit feedback. \emph{ICML}, 814--823.

\bibitem[{Swaminathan \protect\BIBand{} Joachims(2015{\natexlab{b}})}]{swaminathan2015self}
Swaminathan A, Joachims T (2015{\natexlab{b}}) The self-normalized estimator for counterfactual learning. \emph{NeurIPS}.

\bibitem[{Wilder et~al.(2020)Wilder, Horvitz, \protect\BIBand{} Kamar}]{wilder2020learning}
Wilder B, Horvitz E, Kamar E (2020) Learning to complement humans. \emph{arXiv} .

\bibitem[{Yan et~al.(2018)Yan, Chaudhuri, \protect\BIBand{} Javidi}]{yan2018active}
Yan S, Chaudhuri K, Javidi T (2018) Active learning with logged data. \emph{International Conference on Machine Learning}, 5521--5530 (PMLR).

\bibitem[{Yao et~al.(2022)Yao, Chen, Yang, \protect\BIBand{} Narasimhan}]{yao2022webshop}
Yao S, Chen H, Yang J, Narasimhan K (2022) Webshop: Towards scalable real-world web interaction with grounded language agents. \emph{Advances in Neural Information Processing Systems} 35:20744--20757.

\bibitem[{Yoon et~al.(2018)Yoon, Jordon, \protect\BIBand{} van~der Schaar}]{yoon2018ganite}
Yoon J, Jordon J, van~der Schaar M (2018) Ganite: Estimation of individualized treatment effects using generative adversarial nets. \emph{ICLR}.

\bibitem[{Zhan et~al.(2024)Zhan, Ren, Athey, \protect\BIBand{} Zhou}]{zhan2024policy}
Zhan R, Ren Z, Athey S, Zhou Z (2024) Policy learning with adaptively collected data. \emph{Management Science} 70(8):5270--5297.

\bibitem[{Zhang et~al.(2022)Zhang, Zhang, Lipton, Li, \protect\BIBand{} Xing}]{zhang2022exploring}
Zhang YF, Zhang H, Lipton ZC, Li LE, Xing EP (2022) Exploring transformer backbones for heterogeneous treatment effect estimation. \emph{arXiv preprint arXiv:2202.01336} .

\bibitem[{Zhou et~al.(2023)Zhou, Athey, \protect\BIBand{} Wager}]{zhou2023offline}
Zhou Z, Athey S, Wager S (2023) Offline multi-action policy learning: Generalization and optimization. \emph{Operations Research} 71(1):148--183.

\end{thebibliography}
